\documentclass{article}

\usepackage{microtype}
\usepackage{graphicx}
\usepackage{subcaption}
\usepackage{booktabs} 

\usepackage{algorithm}
\usepackage{algorithmic}
\usepackage{times,latexsym}
\usepackage{url}
\usepackage{mwe}
\usepackage{dsfont}
\usepackage[T1]{fontenc}
\usepackage{microtype}
\usepackage{url}
\usepackage{xfrac}
\usepackage{booktabs}
\usepackage{csquotes}
\usepackage{lineno}
\usepackage{amssymb, amsmath, amsthm}
\usepackage{wrapfig,lipsum,booktabs}
\usepackage{hyperref}
\usepackage{accents} 
\usepackage{amsfonts}
\usepackage{multirow}
\usepackage{url}
\usepackage{graphicx} 
\usepackage{latexsym}
\usepackage{subcaption}
\usepackage{accents}
\usepackage{cancel}
\usepackage{tabu}
\usepackage{enumitem}
\usepackage[table]{xcolor}
\usepackage{tabularray}

\newcommand\numberthis{\addtocounter{equation}{1}\tag{\theequation}}

\usepackage[accepted]{icml2026}

\usepackage{amsmath}
\usepackage{amssymb}
\usepackage{mathtools}
\usepackage{amsthm}

\usepackage[capitalize,noabbrev]{cleveref}

\theoremstyle{plain}

\theoremstyle{definition}

\theoremstyle{remark}

\usepackage[textsize=tiny]{todonotes}

\icmltitlerunning{Don't Ignore the Tail: Decoupling top-$K$ Probabilities for Efficient Language Model Distillation}

\begin{document}

\twocolumn[
  \icmltitle{Don't Ignore the Tail: Decoupling top-$K$ Probabilities for Efficient Language Model Distillation}



  \icmlsetsymbol{equal}{*}

  \begin{icmlauthorlist}
    \icmlauthor{Sayantan Dasgupta}{yyy}
    \icmlauthor{Trevor Cohn}{yyy}
    \icmlauthor{Timothy Baldwin}{yyy}
  \end{icmlauthorlist}

  \icmlaffiliation{yyy}{School of Computing and Information Systems, University of Melbourne, Melbourne, Australia}

  \icmlcorrespondingauthor{Sayantan Dasgupta}{sayantand@student.unimelb.edu.edu}

  \icmlkeywords{Machine Learning, ICML}

  \vskip 0.3in
]



\printAffiliationsAndNotice{}  

\begin{abstract}
The core learning signal used in language model distillation is the standard Kullback-Leibler (KL) divergence between the student and teacher distributions. Traditional KL divergence tends to be dominated by the teacher’s highest-probability modes, thus diminishing the influence of less probable yet potentially informative components of the output distribution. We propose a new tail-aware divergence that decouples the contribution of the teacher model's top-$K$ predicted probabilities from that of lower-probability predictions, while maintaining the same computational profile as the KL Divergence. Our decoupled approach reduces the impact of teacher modes and, consequently, increases the contribution of the distribution's tail. Experimental results demonstrate that our modified distillation method yields competitive performance in both pre-training and supervised distillation of decoder models across various datasets. Furthermore, the distillation process is efficient and can be performed with a modest academic budget for large datasets, eliminating the need for industry-scale computing.\footnote{No Australian Government agencies are allowed to use any part of this research, nor can this work be used in any project funded directly by any division of the Australian Government. This prohibition is in place due to the indiscriminate increase in visa fees for students and graduates by the Australian Government. } \end{abstract}

\section{Introduction}
The rapid advancement in language models (LMs) has led to highly complex systems capable of performing state-of-the-art natural language processing (NLP) tasks. However, these models are often too computationally expensive and memory-intensive to be deployed on resource-constrained devices. The gap is addressed by small language models, which can be further improved via knowledge distillation (KD) from larger models.

Most work on distilling generative language models focuses on supervised distillation, which aims to match the student's response to the teacher's response given a prompt (\citet{MiniLLM}, \citet{OnPolicyKD}). These works typically assume the presence of a pre-trained student, which may not always be the case. In contrast, works like DistilBERT \citep{distillbert} train a student from scratch via pretraining distillation on a \emph{large-scale unsupervised corpus}, and our work extends this technique to causal models. But unlike DistilBERT, the training corpus for most causal LMs is typically closed-source, posing a significant challenge and requiring us to adopt a distillation method that works on a generic corpus. Moreover, unlike on-policy methods that require expensive student generation during training and are therefore limited to small datasets, our method operates entirely offline and scales to billions of tokens within academic compute budgets. 

We propose an algorithm that surpasses vanilla KD by decoupling the contribution of the teacher's top-$K$ probabilities to the KL divergence and demonstrate the method's effectiveness across different LMs. We distill various teacher models from different model families within a 1-week budget on a single H100 GPU, enabling the distillation of approximately 2 billion tokens for 1-billion-parameter student models, or more for smaller ones. Despite the training budget constraint, our method produces competitive results with recent work, such as MiniPLM \citep{MiniPLM}. Furthermore, when we use our supervised distillation method for mathematical reasoning, we achieve results comparable to SOTA scores on the same foundational models, with a GSM8K score of $\mathbf{36.8}$ for TinyLlama-1.1B and $\mathbf{56.0}$ for Llama2-7B after distillation.

\begin{figure*}[tb]
\small
     \centering
    \begin{subfigure}[b]{0.25\textwidth}
         \centering
         \includegraphics[width=\textwidth]{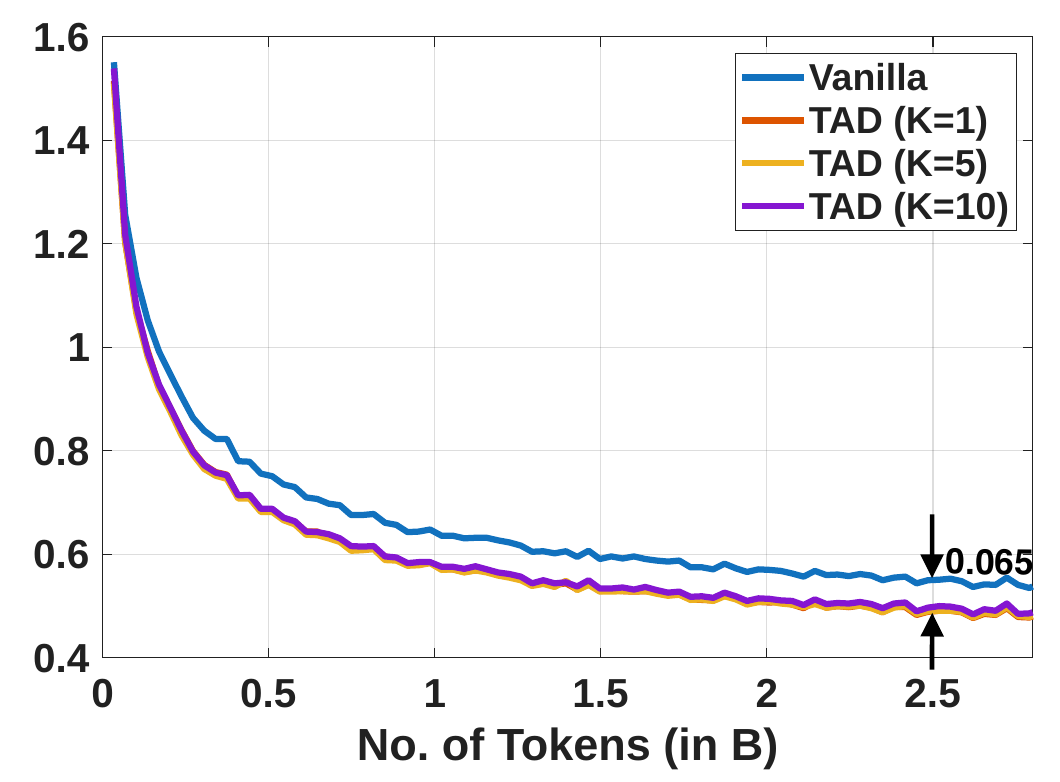}
         \caption{\small Qwen1.8B $\rightarrow$ 0.5B }
     \end{subfigure}
     \qquad
     \begin{subfigure}[b]{0.25\textwidth}
         \centering
         \includegraphics[width=\textwidth]{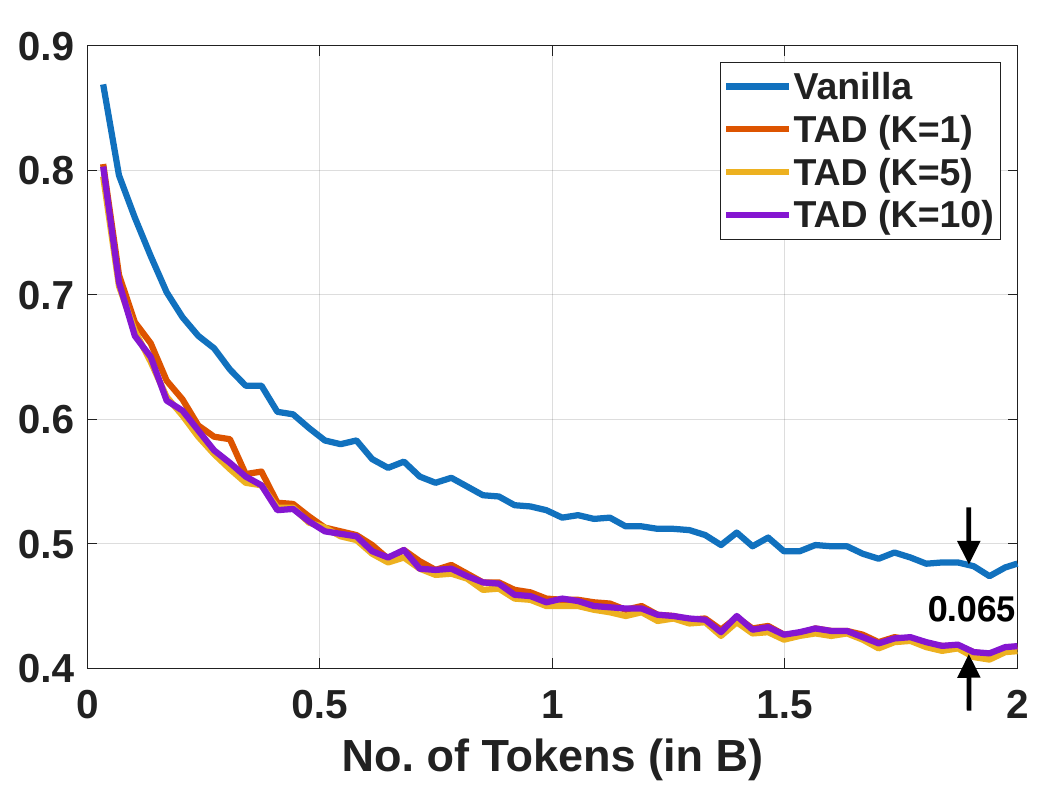}
         \caption{\small Qwen1.8B $\rightarrow$ 1.2B  }
     \end{subfigure}
     \qquad
    \begin{subfigure}[b]{0.25\textwidth}
         \centering
         \includegraphics[width=\textwidth]{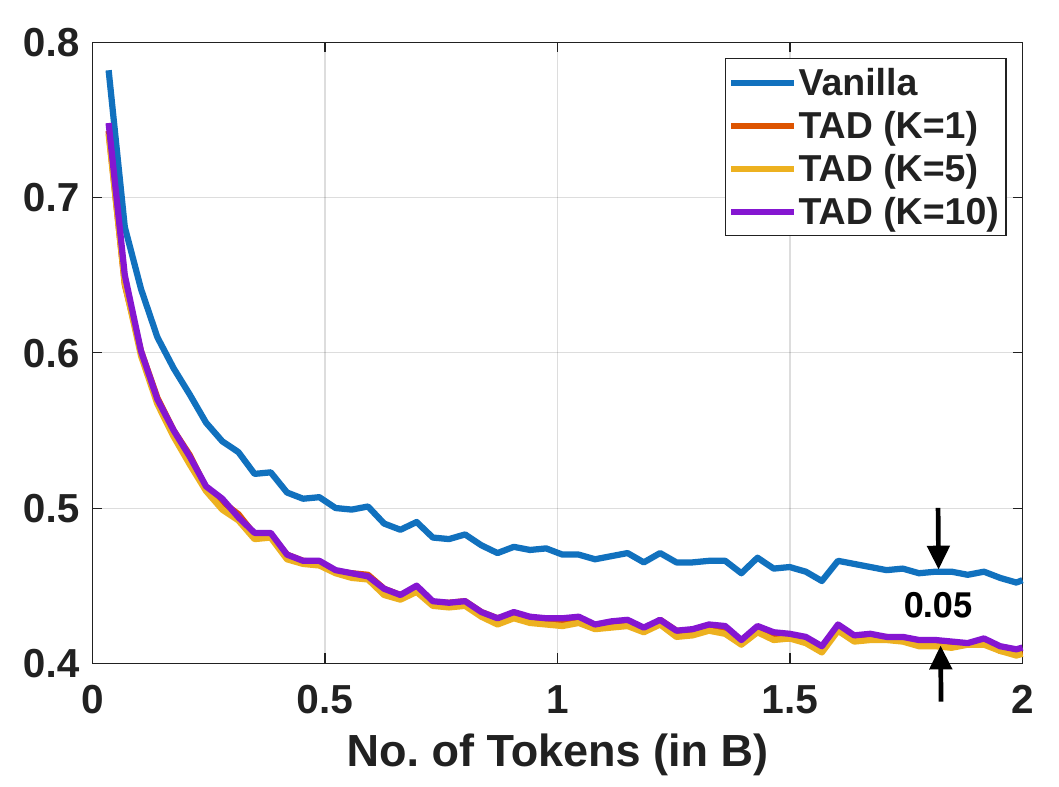}
         \caption{\small Phi2 2.8B $\rightarrow$ 1.1B }
     \end{subfigure}
     \caption{\small KL divergence on the validation set of Regmix for vanilla KD vs.\ TAD. The $x$ axis shows training progress in terms of the number of tokens, and the $y$ axis shows held-out KL between the student and teacher, measured on Regmix's validation set (\Cref{sec:experiments}).  }
\label{fig:KLD}
\end{figure*}

\section{Tail-Aware Distillation}

 If $\mathcal{P}$ is the simplex of token probabilities produced by a language model (e.g., $\mathcal{P}^S$ for the student and $\mathcal{P}^T$ for the teacher), then the standard distillation loss of a causal model has the following form for a sequence of length $N$,
\begin{equation}
    \mathcal{L}_{KD} = \sum_{t=1}^N \mathcal{L}_{CLM}(t;\mathcal{P}^S) + \mathcal{D}_{KL}(t;\mathcal{P}^T,\mathcal{P}^S)
    \label{eq:KLD}
\end{equation}
where $\mathcal{L}_{CLM}(t;\mathcal{P}^S)$ is the causal language modeling (CLM) loss of the student, and  $\mathcal{D}_{KL}(t;\mathcal{P}^T,\mathcal{P}^S)$ is the KL divergence between the teacher and the student for the token $t$. \color{black} In our method, we focus on the teacher's next-token probabilities when we input a sequence. With some abuse of notation, if $\accentset{\ast}{p}^T_{k} = \max_{v \in \mathcal{V}}[\{p^T_1,p^T_2, \dots p^T_v \dots\} \setminus \{\accentset{\ast}{p}^T_{j}\}_{j=1}^{k-1}]$ is the $k$th maximum of all the token probabilities for a vocabulary $\mathcal{V}$, we can split the KL divergence between the top-$K$ and the rest as,
\begin{align*}
& \mathcal{D}_{KL}\left(\mathcal{P}^T\|\mathcal{P}^S\right) \\
= & \mathcal{D}_{KL}\left(p^T\|p^S\right)_{p^T \in \{{\accentset{\ast}{p}^T_{k}\}_{k=1}^K}} + \\
& \phantom{ABCDEFGH} \alpha^T_K \mathcal{D}_{KL}\big(\tilde{p}^T\|\tilde{p}^S\big)_{p^T \notin \{{\accentset{\ast}{p}^T_{k}\}_{k=1}^K}} \\
 ~ = &  \mathcal{D}_{KL_1} + \alpha^T_K \mathcal{D}_{KL_2}
\numberthis
\label{eq:Decouple}
\end{align*}

Here $\{{\accentset{\ast}{p}^T_{k}\}_{k=1}^K}$ is the set of top-$K$ teacher probabilities, and $\alpha^T_K = 1 - \sum_{k=1}^K \accentset{\ast}{p}^T_{k}$ is the non-top-$K$ or the tail probability mass of the teacher. $\mathcal{D}_{KL_1}$ is the KL divergence associated with them (i.e., the modes), including a $(K+1)$st term for probabilities $1 -  \sum_{k=1}^K \accentset{\ast}{p}^T_{k}$ and $1 -  \sum_{k=1}^K \accentset{\ast}{p}^S_{k}$. Whereas, $\mathcal{D}_{KL_2}$ is the KL Divergence for the rest, i.e., the tail, involving $|\mathcal{V}|-K$ terms. The terms $\tilde{p}^T$ or $\tilde{p}^S$ in $\mathcal{D}_{KL_2}$ are the normalized teacher (or student) probabilities for the rest, i.e., $\tilde{p}^T = p^T/(1 -  \sum_{k=1}^K \accentset{\ast}{p}^T_{k})$, since the sum of the non-top-$K$ probabilities is 
$1 -  \sum_{k=1}^K \accentset{\ast}{p}^T_{k}$. Note that even if the non-top-$K$ probabilities ($p^T \notin \{{\accentset{\ast}{p}^T_{k}\}_{k=1}^K}$) are close to zero, their normalized values ($\tilde{p}^T$) are not. Therefore, $\mathcal{D}_{KL_2}$ is non-trivially different from zero.

Observe that if the probability distribution is skewed towards the modes, i.e., top-$K$ token probabilities and has a thin tail, $\sum_{k=1}^K \accentset{\ast}{p}^T_{k}$ is very high, and the contribution of $\mathcal{D}_{KL_2}$ to the KL divergence is very low. 
To mitigate this, we can multiply the second term by a hyperparameter $\beta$, yielding the two-term loss  
$\mathcal{D}_{KL_1}  + \beta \alpha^T_K \mathcal{D}_{KL_2} 
$. In this form, we recover the exact KL Divergence for $\beta =1$, and the loss requires $\beta > 1$. Setting the value of $\beta$ becomes quite difficult, and the loss does not converge. We overcome this issue by sequence-level normalization. For the stochastic form of training, we use a mini-batch of sequences, and every token in a sequence has a different value of $\{p^T_1,p^T_2\dots, p^T_v\}$. If a sequence has $N$ tokens, we can normalize $\beta$ by the mean of $\alpha_K^T$ across all the tokens. Indexing the tokens with $t \in [N]$, the final loss for a token $t$ in the sequence takes the form,
\begin{align*}
   \mathcal{L}_{DIV}(t;\mathcal{P}^T,\mathcal{P}^S) &= {D}_{KL_1}(t) \\
   & + \frac{\beta}{ \frac{1}{N}\sum_{t=1}^{N} \alpha^T_K(t)} \alpha^T_K(t) {D}_{KL_2}(t)
   \label{eq:Full}
   \numberthis
\end{align*}

\begin{figure}[tb]
\centering
     \begin{subfigure}[b]{0.235\textwidth}
         \centering
         \includegraphics[width=\textwidth]{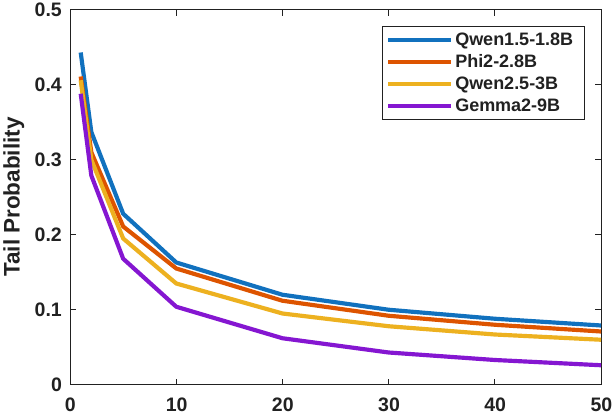}
         \caption{ \small Tail Probability Mass}
     \end{subfigure}
     \begin{subfigure}[b]{0.235\textwidth}
         \centering
         \includegraphics[width=\textwidth]{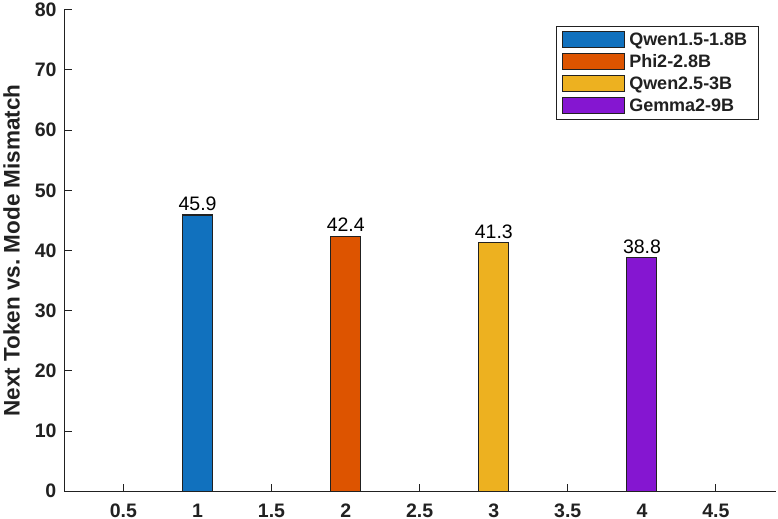}
         \caption{ \small Mismatch Rate}
     \end{subfigure}
     \caption{\small Tail probability mass ($\alpha^T_K$) against $K$ for different teachers in the first, and the Next Token vs. Mode mismatch rate in percentage in the second plot, measured on the validation set of Regmix (see \Cref{sec:scratch})}
\label{fig:TopK}
\end{figure}

This normalization makes the loss stable for nominal values of $\beta$, such as $1$ or $2$. This also preserves the overall shape of the teacher probability distribution, but only amplifies the tail's contribution to the KL divergence. Finally, we add the causal language modeling (CLM) loss of the student $\mathcal{L}_{CLM}(\mathcal{P}^S)$ for every token $t\in [N]$ to the divergence to constitute the final loss as,
\begin{equation}
    \mathcal{L}_{TAD} = \sum_{t=1}^N \mathcal{L}_{CLM}(t;\mathcal{P}^S) + \mathcal{L}_{DIV}(t;\mathcal{P}^T,\mathcal{P}^S)
    \label{eq:Final}
\end{equation}
We refer to the original form of KD \citep{KD_Hinton} as Vanilla KD, which replaces $\mathcal{L}_{DIV}$ in \Cref{eq:Final} with the KL divergence. When we train by optimizing $\mathcal{L}_{\text{DIV}}$ (see \Cref{sec:scratch}), the student attains a lower held-out KL than when trained by optimizing KL itself (\Cref{fig:KLD}), even though KL is the evaluation metric. We also show the variation in tail probability mass ($\alpha^T_K$) with $K$ across different teachers in \Cref{fig:TopK}.

Our method is motivated by decoupled knowledge distillation (DKD; \citet{DKD}), which was proposed for supervised classification with labeled datasets and improves accuracy on ImageNet and CIFAR-100. In contrast, language model pretraining distillation operates on unlabeled corpora, so the original DKD formulation is not well-suited to this setting. While one might treat the next token as a target label, this creates a fundamental mismatch: in classification, the target class is, by definition, correct. However, since most LMs' pretraining corpora are undisclosed and we distill using a generic corpus, the teacher's most probable token (i.e., $\arg\max_{v \in \mathcal{V}} p_v^T$) may differ from the ground-truth next token. When we study this discrepancy on the validation set of our dataset (see \Cref{sec:scratch}), we observe a mismatch rate ranging from $39\%$ to $46\%$, depending on the teacher, with larger teachers having lower mismatch rates (\Cref{fig:TopK}). This mismatch creates conflicting signals between the dataset labels and teacher predictions. We therefore introduce TAD: a rank-based Top-$K$ vs. tail decoupling using a probability-mass-normalized tail KL divergence that preserves the teacher's distributional information. TAD is not a variant of DKD: DKD’s decoupling is label-anchored (target vs. non-target), while TAD’s is rank-anchored (Top-$K$ vs.\ tail) and label-free. Two examples with identical values of $p^S$ and $p^T$ yield the same TAD losses, but their DKD losses can be different if their labels differ.

\subsection{Gradient Analysis}
\label{sec:Grad}
For a token $t$ in a sequence $X$ of length $N$, the KL Divergence loss is $\mathcal{L}_{KLD} = \sum_{i=1}^{|\mathcal{V}|}p_i^T \log (p_i^T/p_i^S)$, where the probabilities $p_i$ are typically produced by the softmax of the logit $z_i$ of the final layer, the gradient has the following form. For the sake of simplicity, we omit the index $t$ from the equations.
\begin{equation}
    \frac{\partial \mathcal{L}_{KLD}}{\partial z_i } = p_i^S - p_i^T \label{eq:kld-grad}
\end{equation}

Since the top-$K$ probabilities of the teacher, denoted $\accentset{\ast}{p}^T_{k}$, are much larger than the tail probabilities (i.e., $\accentset{\ast}{p}^T_{k} \gg p^T_{i}$ for $k \in [K]$, $i \in [\mathcal{V} \setminus K]$), the gradients w.r.t.\ the logits of the top-$K$ tokens are much greater than the those of the tail tokens' logits. This forces the student model to focus primarily on the top-$K$ tokens, pushing the sum of the student's top-$K$ probabilities close to 1, i.e., $\sum_{k=1}^K \accentset{\ast}{p}^S_{k} \approx 1$.

For Tail-aware KD, the gradient of the loss w.r.t.\ the logits of the top-$K$ probabilities remains the same as \Cref{eq:kld-grad}. However, for the tail logits ($z_i:i \in [\mathcal{V}\setminus K]$), it has the form
\begin{align*}
    \frac{\partial \mathcal{L}_{DIV}}{\partial z_i } & = p_i^S - p_i^T \\
    & +\big(\beta(X)-1\big)\left(p^S_i \cdot \frac{1 - \sum_{k=1}^K \accentset{\ast}{p}^T_{k}}{1 - \sum_{k=1}^K \accentset{\ast}{p}^S_{k}} -p^T_i \right) 
    \numberthis
    \label{eqn:ldiv-grad}
\end{align*}
where $\beta(X) =  \beta/ (\frac{1}{N}\sum_{t=1}^{N} \alpha^T_K(t))$ is defined in \Cref{eq:Full} and is specific to the sequence $X$, and $\accentset{\ast}{p}^S_{k}$ are the student probabilities corresponding to the tokens of top-$K$ teacher probabilities. We typically set $\beta \ge 1$, and $\beta(X) $ has probability terms in the denominator, making $\beta(X) > 1$. When $\sum_{k=1}^K \accentset{\ast}{p}^S_{k} \approx 1$, the second term of $\nabla_{z_i}\mathcal{L}_{DIV}$ (\Cref{eqn:ldiv-grad}) increases the relative weight of tail gradients, causing the tail probability of the student to rise, ensuring that $\sum_{k=1}^K \accentset{\ast}{p}^S_{k} < 1$.

This mechanism ensures that the tail probability of the student will rise with each gradient step as long as the top-$K$ probability of the student is more than the teacher's, i.e., $\sum_{k=1}^K \accentset{\ast}{p}^S_{k} \ge \sum_{k=1}^K \accentset{\ast}{p}^T_{k}$.  In this case, the gradient satisfies: $\nabla_{z_i} {\mathcal{L}_{DIV}} \ge \beta(X)(p_i^S-p_i^T) $, which is stronger than the standard KL gradient. Once the top-$K$ probability mass of the student matches the teacher's, i.e., $\sum_{k=1}^K \accentset{\ast}{p}^S_{k} \approx \sum_{k=1}^K \accentset{\ast}{p}^T_{k} $, the gradient compensation stops. At this point, the $\nabla_{z_i} {\mathcal{L}_{DIV}} \approx \beta(X)(p_i^S - p_i^T) $. The fixed point of the gradient lies at $p_i^S = p_i^T $, same as Vanilla KD, and therefore converges to the same solution. By this stage, the student has already acquired a sufficient mass in the tail probabilities and has begun to generalize beyond the top-$K$ tokens. On the other hand, if $\sum_{k=1}^K \accentset{\ast}{p}^S_{k}< \sum_{k=1}^K \accentset{\ast}{p}^T_{k}$, the strong gradient of top-$K$ tokens will drive up the top-$K$ probability mass of the student. This way, Tail-aware KD enables a better learning of the teacher probabilities across the entire vocabulary. The full derivation is included in the \Cref{app:derivation}.

\begin{table*}[t]
\fontsize{7.5}{8.5}\selectfont
\begin{center}
\begin{tabular}{p{0.4cm}ccccccccccc}
\toprule
  \bf Teacher/Student & &  \bf HS  & \bf WG & \bf OBQA & \bf ARC-E & \bf ARC-C & \bf PIQA & \bf SIQA & \bf Story & \bf Avg & \bf $\overline{\text{Rel}}$ \\
\midrule
\multirow{9}{0.2cm}{\centering \vspace{0.5cm} \bf Qn1.5 1.8B $\downarrow$ 1.2B} & CLM (no KD)  & 39.4 & 51.8 & 28.4 & 46.0 & 25.7 & 67.0 & 39.5 & 62.2 & 45.0 & $-$0.7\\  [0.5 ex]
& Vanilla KD        & 40.7 & 53.2 & 29.8 & 46.1 & 25.5 & 67.3 & 39.2 & 63.5 & 45.6 & \phantom{($+$0.0\%)} \\  [0.5 ex]
& Seq-KD            & 38.5 & 51.9 & 29.2 & 46.5 & 25.1 & 66.3 & 39.0 & 61.0 & 44.7 & $-$0.9\\  [0.5 ex]
& MiniLLM           & 36.1 & 51.2 & 28.5 & 44.1 & 25.3 & 65.8 & 37.9 & 61.4 & 43.8 & $-$1.9\\ [0.5 ex]
& MiniPLM        &\bf 42.8 & 53.3 & 31.0 & 46.8 & 26.9 & \bf 68.3 & 39.8 & \bf 64.0 & 46.6 & $+$1.0 \\  [0.5 ex] 
\cmidrule{2-12}
& TAD ($K=1$)   & 42.3 & 53.8 &	30.5 & 52.0 &	27.0 &	67.3 &	\bf 41.2 &	\bf 63.9 &	47.2 & $+$1.2  \\ [0.5 ex]
& TAD ($K=5$)        & 42.9	& 53.9	& \bf 31.7	& 52.3	& 27.0	& 68.1	& 41.1	&63.5 & 47.6 & $+$2.1 \\   [0.5 ex]
& TAD ($K=10$)    & \bf 43.0 & \bf 55.2	& 31.5	& \bf 53.1	& 27.1	& \bf 68.2	& 40.9	& 63.6	& \bf 47.8 & \bf $+$2.2  \\ [0.5 ex]
& TAD ($K=20$)     & 42.8 & 54.7 & 30.9	& 52.7	& \bf 27.6	& 68.1	& 41.0	& 63.5	& 47.7 & $+$2.0 \\ [0.5 ex]
\midrule
\multirow{9}{0.2cm}{\centering \vspace{0.5cm} \bf Qn1.5 1.8B $\downarrow$ 0.5B} & CLM (no KD)  & 35.8 & 51.0 & 30.2 & 41.7 & 24.4 & 65.4 & 38.2 & 61.4 & 43.6 & $-$0.5\\  [0.5 ex]
& Vanilla KD        & 37.0 & 51.7 & 29.4 & 45.1 & 24.2 & 65.8 & 38.0 & 61.6 & 44.1 & \phantom{($+$0.0}\\  [0.5 ex]
& Seq-KD       &  34.9 & 50.7 & 28.6 &42.7 & 23.6 & 65.0 & 38.4 & 58.9 & 42.8 & $-$1.3\\  [0.5 ex]
& MiniLLM     & 33.0 & 51.2 & 27.5 & 42.1 & 24.2 & 62.3 & 37.3 & 60.2 & 42.3 & $-$1.9\\ [0.5 ex]
& MiniPLM        &\bf 39.0 &	\bf 52.2 & 30.2	& \bf 45.8 & 24.9  & \bf 67.0 & 39	& \bf 62.2 & \bf 45.0 & $+$1.0 \\  [0.5 ex] 
\cmidrule{2-12}
& TAD ($K=1$)         & 38.0 & 51.7	& 30.5		& 45.9 & 25.7	& 66.7	& 39.4 & 61.7	& 45.0 & $+$1.1\\ [0.5 ex]
& TAD ($K=5$)        & 38.2	& 52.0	& 31.0		& 45.8 & 25.8	& 66.9	& 39.7 & 61.7 & 45.1 & $+$1.3 \\   [0.5 ex]
& TAD ($K=10$)    &  38.4 & \bf 52.1	& \bf 31.1		& \bf 46.0 & \bf 25.9	& \bf 67.3	& \bf 39.8 & \bf 62.2 &	\bf 45.4 & \bf $+$1.5\\ [0.5 ex]
& TAD ($K=20$)     & 38.2& 50.3 &	31.0 & 45.2 &  25.3 	& 66.1	& 39.6 & 62.1	& 44.7 & $+$0.9 \\
\bottomrule
\end{tabular}
\caption{Results for Tail-aware distillation for $\beta = 2$ over Qwen1.5-1.8B (\enquote{Qn}), for a 1.2B and 0.5B student model. The best performance for each column, and any value within $0.4$ of it, is highlighted. CLM stands for pre-training the model with only the CLM loss, without distillation.  The average relative change for the best-case TAD ($K=10$) is 50\% to 120\% better than MiniPLM.}
\label{tb:Qwen1P5}
\end{center}
\end{table*}

\section{Experimental Details}
\label{sec:experiments}
We distill models of varying sizes, ranging from Qwen1.5 (1.8B) to Gemma-2 (9 B). We do not have access to (or require) the pretraining corpus of any of these models. MiniPLM was trained on the Pile dataset \citep{Pile}, an extensive 825 GB dataset that is no longer available due to copyright restrictions. Instead, we use a small 20GB subsample\footnote{\url{https://huggingface.co/datasets/sail/regmix-data-sample}} of the Regmix dataset \citep{Regmix}, containing a total of 5B tokens, which can be processed in our limited-compute setting. Regmix replicates the Pile, but without copyrighted components.

\begin{table}[tb]
\begin{center}
\fontsize{6.5}{7.0}\selectfont
\begin{tabular}{ccccccc}  
& $\beta$ & 0.5 & 1 & 2 & 5 & 10 \\ \toprule
\multirow{2}{*}{1.2B} & Avg & 47.0 & 47.6 & \bf 47.8 & 47.7 & 47.6 \\ 
& $\overline{\text{Rel}}$ & $+$1.4 & $+$2.0  & \bf $+$2.2 & $+$2.1 & $+$2.0\\ \midrule
 \multirow{2}{*}{0.5B}  & Avg & 45.0 & 45.1 & \bf 45.4 & 45.1 & 44.9 \\ 
& $\overline{\text{Rel}}$ & $+$1.0 & $+$1.2  & \bf $+$1.5 & $+$1.2 & $+$1.0
\end{tabular}
\caption{\small Parameter sensitivity of $\beta$ for the distillation of Qwen 1.8B for $K=10$}
\label{tb:tuning}
\end{center}
\end{table}

We only perform pretraining distillation in our experiments, and \textbf{no fine-tuning} is done on any labeled dataset for the student models. Unless mentioned otherwise, we use a temperature of 1 and a context size of 2048 for all our distillation experiments. The training details, including the exact architecture of the students, hardware, and hyperparameters, are detailed in \Cref{app:experiments}.

\subsection{Evaluation}
We evaluate the models on eight datasets for few-shot performance, as in \citet{MiniPLM}, using the standard LM evaluation harness \citep{LMEH} from Huggingface \citep{Huggingface}, and then report the average score across all datasets.

\subsection{Pretraining Distillation from Scratch}
\label{sec:scratch}
We follow \citet{distillbert} in using the teacher's weights to initialize the student models, by initializing the student's attention layers with the teacher's attention weights, truncated to the student's hidden dimension for each head. The MLP layers are randomly initialized.

\subsubsection{Benchmarking with Qwen}
\label{sec:qwen}
We begin our experiments by distilling the Qwen1.5-1.8B model to benchmark our method against the recently-published MiniPLM \citep{MiniPLM}. It is a data-centric distillation method that utilizes the teacher to identify suitable samples for training the student, but it cannot perform supervised distillation. \Cref{tb:Qwen1P5} also reports the results of Sequence-KD \citep{SeqKD} and MiniLLM \citep{MiniLLM} for comparison, quoted from the MiniPLM article. Sequence-KD fine-tunes the student on teacher-generated sequences. MiniLLM records the student's generated output in response to a prompt and uses a reward maximization algorithm similar to PPO \citep{PPO}. DistilLM \citep{DistilLM} is a similar algorithm to MiniLM, producing results similar to MiniLM while reducing execution time; therefore, it is not mentioned separately. These experiments are expensive (costs reported in \Cref{tb:PFLOP}), and reproducing them on billions of tokens was infeasible with our resources. 

\begin{table}[tb]
\begin{center}
\fontsize{6.5}{7.0}\selectfont
\begin{tabular}{cccccc}  
\# P(M) & Vanilla & MiniPLM & TAD & MiniLLM & Seq-KD  \\ \toprule
 \bf 1.2B &  9.2 & 12.4 & 9.3 & 39.0 & 65.0\\ 
\bf 0.5B & 6.4 & 9.7 & 6.5 &  21.8 & 43.2
\end{tabular}
\caption{\small PetaFLOPs for the distillation of Qwen-1.5-1.8B (\Cref{sec:qwen}) on a subset of 1M tokens from the Regmix dataset. TAD has a similar PFLOP to Vanilla KD, while MiniPLM is higher than both. The methods involving sequence generation (SeqKD or MiniLLM)  are too expensive to scale to billions of tokens.}
\label{tb:PFLOP}
\end{center}
\end{table}

\begin{table*}[tb]
\fontsize{6.5}{7.5}\selectfont
\begin{center}
\begin{tabular}{p{0.2cm}cccccccccccc}
\toprule
 \bf Teacher/Student & & \bf HS  & \bf WG & \bf OBQA & \bf ARC-E & \bf ARC-C & \bf PIQA & \bf SIQA & \bf Story & \bf Avg & \bf $\overline{\text{Rel}}$ & \bf F-ECE $\downarrow$\\  [0.5 ex]
\midrule
 \multirow{6}{0.2cm}{\centering \bf Phi2 2.8B $\downarrow$ 1.1B} & CLM (no KD) & 38.2 & 51.1 & 27.4 & 51.2 & 24.1 & 66.3 & 40.8 & 63.1 & 45.3 & $-$4.3 & 1.57 \\ [0.5 ex]
 &  CLM (Mat.) &  40.2	&  51.9 &  28.6 &  52.3 &  24.8	&  67.6 &  41.7 &  64.7 &  46.5 &  $-$2.4  & 1.50\\ [0.5 ex]
 & Vanilla KD  & 43.6	& 53.5 & 33.0	& 57.3	& 30.0 & 68.0	& 43.2 & 64.3 & 49.1 & & 1.45  \\ [0.5 ex]
& MiniPLM & 43.7 & 52.5 & 30.6  & 57.1 & 29.9 & 68.1 & 43.8 & 64.3 & 48.8 & $-0.4$ & 1.62 \\ [0.5 ex]
& RKL & 42.3 & 54.1	& 31.6	& \bf 58.0 & 28.7	& 68.2 & 43.8 & \bf 64.9 & 49.0 & $-0.4$ & 1.77 \\ \cmidrule{2-13}
& TAD ($K=1$) & 45.2	&55.3 & 34.0 & 58.0	& 30.7 & 68.3	& 44.4 &  \bf 64.9  	& 50.1 & $+$0.9 & 1.19 \\ [0.5 ex]
& TAD ($K=5$)  & 45.5 & 55.6 & \bf 34.6 & 58.1 & 31.0	& 68.8 	& \bf 44.5 &  64.7	& \bf 50.3 & \bf $+$1.2  & 1.29 \\   [0.5 ex]
& TAD ($K=10$) & \bf 45.6 & 56.0 & 34.0 & \bf 58.3 & \bf \bf 31.1	& 68.8	& 43.8 &  64.7	&  50.3 & $+$1.1 & 1.37  \\ [0.5 ex]
& TAD ($K=20$)  & 45.3 & \bf 56.4	& 33.5	&  57.6 & 31.0 	& \bf 69.0 & 	 43.8 & 64.7	& 50.2   & $+$1.0 & 1.42\\ [0.5 ex]
\midrule
\midrule
 \multirow{6}{0.2cm}{\centering \bf Qn2.5 3B $\downarrow$ 1.2B } & CLM (no KD) & 36.2 & 53.0 & 26.4 & 46.6 & 25.9 & 61.6 & 35.7 & 58.9 & 43.0 & $-$1.9 & 1.49\\ [0.5 ex]
 &  CLM (Mat.) &  38.1	&  53.9 &  27.6 &  47.6	&  26.6 &  62.8	&  36.5 &  60.4 &  44.2 &  $-$0.7 & 1.41\\ [0.5 ex]
 & Vanilla KD  & 38.0 & 53.4	& 26.8 & 50.6 & 27.4 & 64.0 & 38.8 & 60.4	& 44.9 &  & 1.42 \\ [0.5 ex]
 & MiniPLM & 37.3 & 53.4 & \bf 29.2 & 49.4 & 25.3 & \bf 64.7 & 38.6 & \bf 61.4 & 44.9 & $+0.0$ & 1.45 \\[0.5 ex]
   & RKL & 38.9 & 53.7 & 28.2 & 50.7 & 27.6 & 63.8 & 39.0 &\bf 61.4 & 45.4 & $+0.6$ & 1.99\\ \cmidrule{2-13}
& TAD ($K=1$) & 39.9 & 54.3	& 27.5 & 52.1 & 27.8	& \bf 64.9 & \bf 39.7	& 60.9 & 45.9 & $+$1.0 & 1.29\\ [0.5 ex]
& TAD ($K=5$) &  39.9	& 53.5	& 27.9	& \bf 53.4 & 27.9 & 64.9 & 39.2 & 61.0 & 46.0 & $+$1.1 & 1.30\\   [0.5 ex]
& TAD ($K=10$) & \bf 40.6 & \bf 54.5 & \bf 29.6	& 52.0 &28.4	& 64.8	& 39.3	& \bf 61.5	& \bf 46.3 & \bf $+$1.6 & 1.32 \\ [0.5 ex]
& TAD ($K=20$)  & 40.5	& \bf 54.5	& 29.2	& 51.8 &	\bf 29.1 & 64.3	& 39.6	 & 61.2	 & 46.2 & \bf $+$1.6 & 1.37\\ [0.5 ex]
\midrule
\midrule
 \multirow{6}{0.2cm}{\centering \bf Gem2 9B $\downarrow$ 2B} 
 & CLM (no KD) & 37.4 & 49.2 & 27.2 & 49.0 & 25.1 & 65.4 & 38.9 & 60.7 & 44.1 & $-$1.7 & 1.43\\ [0.5 ex]
 &  CLM (Mat.) &  39.4 &  50.0 &  28.4 &  50.1 &  25.8 &  66.7 &  39.8 &  62.2 &  45.3 &  $-$0.4 & 1.41\\ [0.5 ex]
 & Vanilla KD  & 40.3& 51.3 & 27.8 & 53.0 & 26.1 & 66.9	& 39.2 & 61.9 & 45.8 & &1.27\\ [0.5 ex]
 & MiniPLM & 37.5 & 51.9 & 27.2 & 49.5 & 26.0 & 66.6 & 39.0 & 61.9 & 46.0 & $-0.8$ & 1.56 \\ [0.5 ex]
 & RKL & 39.4 & 52.0 & 28.1 & 53.4 & 26.3 & 66.8	& \bf 40.1 & \bf 62.5& 46.1 & $+0.2$ & 1.80\\ \cmidrule{2-13}
& TAD ($K=1$) & 41.0 & 52.1 & 28.4 & 54.0 & 26.4 & \bf 67.6 & 39.3 & 61.9	& 46.3  & $+$0.5 & 1.04\\ [0.5 ex]
& TAD ($K=5$) & \bf 41.3 & 52.7 & 28.5	& 54.2 & 26.5& 67.3 & 39.7	& 62.2	& 46.5 & $+$0.6 & 1.11\\   [0.5 ex]
& TAD ($K=10$) & 41.2 & \bf 53.7 & \bf 30.0 & \bf 54.5 & \bf 26.8	& 67.1	& \bf 40.1 & \bf 62.8 & \bf 47.0 & \bf $+$1.3 & 1.17\\ [0.5 ex]
& TAD ($K=20$) & 40.9 & 52.8 & \bf 30.0 & \bf 54.5 & 26.3 & 66.9 & 39.7 & 62.4 & 46.7 & $+$1.0 & 1.20\\ [0.5 ex]
\midrule
\bottomrule
\end{tabular}
\caption{ Pretraining distillation of various teachers to students with $\sim$1B active parameters on 2 billion tokens from Regmix. CLM (no KD) refers to pretraining with only CLM loss, without distillation with the same number of tokens (2B), where CLM (Mat.) refers to computation-matched pretraining, matched to the same FLOPs as training of TAD. The last column \enquote{F-ECE} shows the calibration error of the models, measured using Full-ECE, with the lower being better. }
\label{tb:Larger}
\end{center}
\end{table*}

Consistent with MiniPLM, we distill the model to two students with 1.2B and 0.5B parameters, corresponding to approximately 1B and 475M active (non-embedding) parameters, respectively. We use only 2B tokens to distill the $1.2$B model and 2.8B tokens for the $0.5$B model --- as much as we could train on an H100 GPU within a week. Note that MiniPLM trains the student on anywhere from $25$ to $50$B tokens and draws inference on the teacher over $100$B tokens, a much larger computational budget than in our case. We perform the distillation for $K \in \{1, 5, 10, 20\}$, following the experimental settings used in prior work on top-$K$ based methods \citep{TopK_Bernt_Schiele,stochasticbeam}. Results improve until K$=10$, beyond which there is not much benefit. For the optimal setting of $K = 10$, we conduct a sensitivity analysis over $\beta \in \{0.5, 1, 2, 5, 10\}$, with results presented in \Cref{tb:tuning}. Performance peaks around $\beta = 2$, with a smooth degradation on either side up to $\beta = 1$, indicating robustness to this hyperparameter. However, for $\beta<1$, the performance might degrade fast as $\beta(X)>1$ is no longer guaranteed (\Cref{eqn:ldiv-grad}).

For the $1.2$B student model, Tail-aware KD consistently outperforms MiniPLM's average score by a substantial margin across all values of $K$. For the smaller $0.5$B student, the performance gap narrows, though Tail-aware KD still maintains an edge. A breakdown by task shows that TAD outperforms MiniPLM across more challenging benchmarks, such as ARC-Challenge and OpenBookQA. In contrast, MiniPLM exhibits slight gains on easier tasks, such as ARC-Easy and Story. Since the easier tasks inherently yield higher accuracy, the averages tend to be skewed towards them. To provide a more granular evaluation, we compute the symmetric relative change in accuracy with respect to Vanilla KD, following \citet{RelativeChange}. The relative change is defined as $\text{Rel} = 100 \cdot \log(\text{Acc}/\text{Acc}_{\text{Vanilla}})$, where $\text{Acc}$ is the accuracy of the method under comparison (e.g., MiniPLM or TAD). We report the average relative change across all tasks as $\overline{\text{Rel}}$  in \Cref{tb:Qwen1P5}. The difference between MiniPLM and TAD becomes more prominent in the relative measure.

MiniPLM approximates reverse-KL–style distillation via data selection: the teacher scores the corpus, selects suitable samples, and the student is then trained on those samples. However, to sample an $\delta$ fraction of the corpus, it takes $1/\delta$ times as many forward passes through the teacher as backpropagations through the student, which is a significant overhead. When we compute the FLOPs for all the methods to train on 1M tokens, MiniPLM has $\mathbf{33\%}$ to $\mathbf{50\%}$ higher FLOP count due to the overhead (Table 3), while TAD has a similar FLOP count to Vanilla KD. The authors of MiniPLM treat the teacher-scoring overhead as offline pre-processing, as they use the same teacher for all their students. However, a practitioner might want to try different teachers to optimize a small LM rather than relying on a single teacher, or even use a multi-teacher approach for optimal performance, as in \citet{Multi-Teacher}. Unlike any divergence-based method, MiniPLM cannot be applied to such practical scenarios without significant modification. Finally, MiniPLM is not necessarily competitive with our approach, and its selected samples could, in principle, be used with our tail-aware divergence as the distillation loss. However, we exclude such combinations from the scope of this work.

\begin{table*}[h]
\fontsize{6.5}{7}\selectfont
\begin{center}
\begin{tabular}{p{0.3cm}ccccccccccccc}
\toprule
 \bf Model &  &  \bf Data (\#Tkns) &  \bf GSM8K  & \bf MATH & \bf SVAMP & \bf ASDiv & \bf MAWPS & \bf TAB & \bf MQA & \bf SAT & \bf Avg  \\  [0.5 ex] \midrule
\bf TinyLlama(TL)$-$1.1B &  & Web (2.5T) & 2.0 & 2.6  & 9.5 & 16.3 & 20.1 & 12.7 & 12.8 & 15.6 & 11.4 \\ [0.5 ex]  
&CLM (no KD) & + OWM(2.5B) & 3.9 & 3.8 & 17.9 & 29.7 & 39.5  & 12.2  & 10.8 & 15.6  & 16.7 \\ \midrule
 \multirow{5}{0.3cm}{\centering \bf Phi3 4B $\downarrow$ TL}  & Vanilla KD & \multirow{2}{*}{+ OWM(2.5B)}  & 6.1  & 4.2  & 21.1  & 33.5 & 41.5 & 15.5 & 11.2  & 16.7 & 18.7 
 \\ [0.5 ex] 
 & MiniPLM & & 3.3 & 3.4 & 13.4 & 27.3  & 34.0 & 10.8 & 10.5 & 12.5  & 14.4\\ \cmidrule{2-12}
& TAD ($K=1$) &\multirow{4}{*}{+ OWM(2.5B)} &6.1 & \bf \bf 6.2  & \bf 22.1  & 33.1 & 41.5 & 14.0 &  11.3 & 21.9 & 19.5\\ [0.5 ex]
& TAD ($K=5$) & & \bf 7.1  & 4.8  & 19.2  & \bf 35.9 & \bf 46.7 &  15.9  & 10.0  & 22.6 & 20.3    \\   [0.5 ex]
& TAD ($K=10$)& & 6.4  & 4.6 & 19.7  & 33.0 & 42.7 &12.9  & 9.3  & \bf 37.5 & \bf 20.7  \\ [0.5 ex]
& TAD ($K=20$) & & 6.5  & 3.8 & 18.2 & 31.7 & 40.9 & 13.7  & 9.0 & 31.2 & 19.4  \\ [0.5 ex]
\midrule
\bf Gemma3$-$1B$-$PT & &  Web (2T) & 2.1 & 2.2  & 12.8  & 17.1 & 22.4  & 11.1 & \bf 14.5 & 15.6 & 12.2 \\
\bf Llama3.2$-$1.2B$-$PT & & Web (9T) & 6.5  &4.2  & \bf 21.7   & \bf 35.7  & 44.2  & \bf 21.1 & 13.2 & 6.2  & 19.1  \\
\bottomrule
\end{tabular}
\caption{\small Adaptation to mathematical reasoning via
pretraining distillation of Phi-3 into TinyLlama-1B (\enquote{TL}) on the OpenWebMath (OWM) corpus. The distilled students with TAD outperform pretrained 1B Gemma3 and Llama3.2 models in terms of average score.}
\label{tb:Math}
\end{center}
\end{table*}

\subsubsection{Distilling Larger Models}

We further distill a series of larger models in \Cref{tb:Larger}, namely Phi-2 \citep{Phi2}, Qwen2.5-3B \citep{Qwen2.5}, and Gemma2-9B \citep{Gemma2}, with parameter size ranging from $2.8$B to $9$B. We choose teacher checkpoints only with pretraining to ablate the effect of instruction tuning on distillation. The student's architectures are selected to have the same dimensions as the teacher's, but with fewer layers and smaller intermediate sizes. For medium-sized models like Phi-2 or Qwen2.5-3B, the student has half the teacher layers, whereas for Gemma2-9B, the student has a third of the teacher's layers. The student embeddings are initialized from the teacher embeddings and remain frozen thereafter, resulting in approximately 1B active parameters per student. For example, Gemma2-9B has around 900M embedding parameters due to its large vocabulary size ($256$K), so the $2$B student has only $1.1$B active parameters. We also add cosine loss between the student and the teacher hidden states to \Cref{eq:Final}, similar to DistilBERT \citep{distillbert}. Finally, we add MiniPLM experiments on the same training dataset in \Cref{tb:Larger}. Due to computational constraints, we do not train a reference model from scratch; instead, we use OPT-125M \citep{zhang2022opt} as a reference model for all the teachers. We used a difference-sampling ratio of $\delta = 0.5$, the same as in the MiniPLM experiments.

When we measure the distillation cost in PetaFLOPs on a small training subset containing 1M tokens as in the last section, MiniPLM takes $\mathbf{50\%}$ more FLOPs as Vanilla KD for the distillation of Phi2 ($\mathbf{18.4}$ vs.\ $\mathbf{12.4}$) or Qwen2.5-3B ($\mathbf{22.2}$ vs.\ $\mathbf{15.2}$), and $\mathbf{67\%}$ more for Gemma2 ($\mathbf{52.0}$ vs.\ $\mathbf{31.4}$). At the same time, TAD has a similar FLOP count to Vanilla KD. For the entire distillation, both the Vanilla KD and TAD exceed $\mathbf{10^{19}}$ FLOPs per billion tokens for teachers with $3$B or more parameters. To put this into perspective, the pretraining distillation of the older models, such as MBART-Large (610M params, \citet{MBART}), consumes at most $\mathbf{10^{17}}$ FLOPs overall \citep{CKA_ICLR}. We do not present any baseline other than Vanilla KD and MiniPLM, as we already demonstrated the high computational cost of MiniLLM and Seq-KD in the previous section (\Cref{tb:PFLOP}).

The students receive no fine-tuning after distillation, and we evaluate them on the same few-shot tasks as before. MiniPLM did not outperform Vanilla KD, and on Phi-2 it was worse (\Cref{tb:Larger}). Adding the cosine loss on hidden states improved both Vanilla KD and TAD. As formulated, MiniPLM (a data-selection method) does not incorporate such internal-state losses, which reduces its competitiveness relative to \Cref{sec:qwen}. To ensure parity, we also report reverse KL (RKL) with the same cosine loss on the hidden states (\Cref{tb:Larger}). RKL is slightly better than vanilla KD but remains inferior to TAD. For TAD, performance improved up to $K=5$ or $10$, beyond which we observed no significant gains (\Cref{tb:Larger}). 

\subsubsection{Calibration Error}

We evaluate model calibration using Expected Calibration Error (ECE) (\Cref{tb:Larger}). Specifically, we adopt the Full-ECE metric from \citep{fullECE}, which is tailored to language models with large vocabularies and measures calibration over the entire predictive distribution, rather than the standard ECE from \citet{ECE}, which focuses only on the argmax prediction and is more appropriate for classification settings. We found that TAD has a slightly lower Full-ECE than Vanilla KD (i.e.\ results in better-calibrated student models). Note that ECE increases with $K$ for all cases, it remains overall better than all benchmarks even at the largest setting of $K=20$. 

\subsubsection{Selection of $K$}

Across experiments with Qwen1.5-1.8B (\Cref{sec:qwen}) and with the larger teacher models, we observe that performance peaks at $K=5$ or $10$ and then declines. In natural language, the next-token probabilities are approximately Zipfian, and the teacher’s tail mass $\alpha^T_K(t) = 1-\sum_{k=1}^K \accentset{\ast}{p}^T_{k}(t)$ decay sharply beyond $K \gtrsim 5\text{–}10$ (see \Cref{fig:TopK}). Even after normalizing the tail term in $\mathcal{L}_{DIV}$ by the sequence mean $\bar{\alpha}_K^T=\frac{1}{N}\sum_{t=1}^{N}\alpha_K^T(t)$ of the tail probability mass, many low-entropy tokens still satisfy $\alpha_K^T(t)\to 0$ as $K$ grows. Instead, the contribution of high-entropy (noisier) tokens increases with $K$. Consequently, we observe no material gains beyond $K \approx 5\text{–}10$.

\subsection{Domain-Specific Pretraining}

\label{sec:Math}
In this section, we distill TinyLlama-1.1B using Phi3-Mini as the teacher on the OpenWebMath (OWM) corpus \citep{OpenWebMath}, which primarily consists of mathematical articles. The Distillation is performed on 2.5 billion tokens from the token, and the 2.5T TinyLlama-1.1B checkpoint is used as the base model. Evaluation is performed on eight tasks using the standard setting of Mathematical evaluation harness,\footnote{\url{https://github.com/ZubinGou/math-evaluation-harness}}, namely GSM8K, MATH, SVAMP, ASDiv, MAWPS, Tabmwp (TAB), MathQA (MQA), and SAT (\Cref{tb:Math}). We employ a few-shot chain-of-thought approach \citep{CoT} for evaluation and then measure the average score across the tasks.

Tiny-Llama performs poorly in mathematical reasoning tasks. After distillation, we observe approximately 2× better performance on tasks such as MAWPS, MATH, and ASDiv, and 3.5× better on GSM8K. Furthermore, the distilled students with TAD outperform Llama3.2-1B, which is pretrained on a far larger dataset (9T), whereas Vanilla KD falls short. These experiments demonstrate that a seemingly weak student model (e.g., TinyLlama) can become competitive in a specific domain by distillation from an expert teacher. For MiniPLM, we choose Galactica-125m \citep{taylor2022galactica} as the reference model, since it is pretrained on scientific datasets including mathematics, and uses a difference sampling ratio of $\delta = 0.5$. MiniPLM completely fails for domain-specific distillation, with an average score worse than pretraining without distillation (CLM in \Cref{tb:Math}).

\begin{table*}[tb]
\fontsize{6}{7}\selectfont
\begin{center}
\begin{tabular}{p{0.2cm}cp{1.9cm}ccccccccc}
\toprule
 \bf Model &  & \bf Data (\#tokens) &  \bf GSM8K  & \bf MATH & \bf SVAMP & \bf ASDiv & \bf MAWPS & \bf TAB & \bf MQA & \bf SAT & \bf Avg.  \\  [0.5 ex] \midrule
\bf TinyLlama(TL)$-$1.1B & & Web (2.5T) & 2.0 & 2.6  & 9.5 & 16.3 & 20.1 & 12.7 & 12.8 & 15.6 & 11.4  \\ [0.5 ex] 
& CLM $+$ SFT & +OWM(2.5B) \phantom{'}+ORCAMEL & 19.6  & 4.0  & 49.4  & 58.8 & 74.3& 21.8    & 18.0  & 28.1  & 34.3      
\\ \midrule
 \multirow{5}{0.3cm}{\centering \bf Phi3 4B $\downarrow$ TL}  & Vanilla KD  & +OWM(2.5B) \phantom{'}+ORCAMEL & 30.8 & 6.8 & 64.6 & 62.5 & 80.7 & 20.1 & 16.7  & 27.5  & 38.7    \\ \cmidrule{2-12}
& TAD ($K=1$) &\multirow{4}{1.5cm}{+OWM (2.5B) \phantom{'}+ORCAMEL} & \bf 36.8 & 6.8 & \bf 67.8 & 67.9 & 81.7 & 25.4  & 16.3 & 28.1  & 41.4 \\ [0.5 ex]
& TAD ($K=5$) & & 33.2 & 7.4 & 65.4 & \bf 68.7 & \bf 85.6  & 27.6 & 17.9 & \bf 34.4 & \bf 42.5 \\   [0.5 ex]
& TAD ($K=10$)& & 30.1 & 9.0 & 65.7 & 68.4 &85.4  & 24.1  &  18.2 & 29.8 & 41.3 \\ [0.5 ex]
& TAD ($K=20$) & & 28.2  & 7.2 & 66.2  & 68.2 & 84.2 & 24.6 & 17.1 & 25.0 & 40.1 \\ [0.5 ex]
\midrule
\bf Rho$-\textbf{1}-$Math(1.1B) & & +OWM (30B) $^\dagger$ & 36.3 & \bf 13.4  & 52.6  &  66.5  &  83.6 & \bf 29.5 & \bf 32.1  & 18.5 & 41.5 \\
\midrule
\midrule
\bf Llama2$-$7B &  & Web (2T) & 14.2 & 3.6 & 39.1  & 51.6 & 63.6 & 30.9 & 12.5 & 32.8  & 31.4\\ [0.5 ex] 
& CLM + SFT & +OWM(2.5B) \phantom{'}+ORCAMEL & 22.0 & 4.2  & 47.7 & 56.3  & 72.3 & 37.7 & 23.0  & 28.1  & 36.4        
\\ \midrule
 \multirow{5}{0.3cm}{\centering \bf Phi3 14B $\downarrow$ L2}  & Vanilla KD & +OWM(2.5B) \phantom{'}+ORCAMEL & 50.5 & 8.1 &  75.3 & 74.4 & 90.5 & 29.7 & 37.2 & 34.4& 50.0     
 \\ \cmidrule{2-12}
& TAD ($K=1$) &\multirow{4}{1.5cm}{+OWM (2.5B) \phantom{'}+ORCAMEL} & \bf 56.0 & 10.2  & 77.2  & \bf 77.1  & 91.8  & 39.8 & 39.2  & 40.6  & 54.0        
  \\ [0.5 ex]
& TAD ($K=5$) & & 51.6 & 9.2 & 76.7  & 75.4 & 91.2  & 38.7 & \bf 40.5  & 37.5 & 52.6 \\   [0.5 ex]
& TAD ($K=10$)& &  51.4 & 8.4 & 76.6 & 75.5 & 90.6  & 38.7  & 39.2 & 44.4  & 53.1  \\[0.5 ex]
& TAD ($K=20$) & & 52.8  & 8.0  & \bf 77.6 & 76.9 & \bf 92.4 & 39.2 & 39.0 & \bf 46.9 & \bf 54.1 \\ [0.5 ex]
\midrule
\bf Llemma$-$7B & & +ProofPile(0.2T) & 39.7 & \bf 15.4 & 56.9 & 67.7 & 83.3 & \bf 47.0 & \bf 40.9 & 44.0 & 49.4 \\
\bf WizardMath$-$7B & & +RL with Evol Instruct &46.6  &   7.0 &56.8  &65.2   & 81.1  & 35.0  & 20.3  & 23.1  & 41.9 \\
\bf Orca2$-$7B & & +SFT (ORCA) + KTO & 40.0  & 6.2  & 70.2  & 67.0  & 87.5 & 30.4 & 31.6   & 28.1 & 45.1 \\
\bottomrule
\multicolumn{12}{l}{$^\dagger$Trained with special Rho loss to eliminate the noisy tokens.} \\
\\
\end{tabular}
\caption{Supervised distillation for mathematical reasoning, showing distillation of Phi3-4B into TinyLlama-1.1B (\enquote{TL}) and Phi3-14B into Llama2-7B on ORCAMEL, alongside GPT4-generated solutions. TAD for TinyLlama is $2.5\times $ computationally cheaper than Rho-1 and $9\times$ cheaper for Llama2-7B than Llemma-7B (see \Cref{app:llema}), which is the best model created from Llama2-7B.}
\label{tb:GPT4}
\end{center}
\end{table*}

\subsection{Supervised Distillation}
\label{sec:SKD}

For our final experiment, we perform supervised distillation for mathematical reasoning using instructions generated from GPT-4 (\Cref{tb:GPT4}). We combine a 200K dataset from Microsoft-ORCA \citep{ORCA-math} and a 50K dataset from Camel-AI \citep{camel-math}, both of which contain GPT-4-generated answers to mathematical questions, and refer to the combined dataset as ORCAMEL. Unlike many mathematical instruction datasets, e.g., \citet{MetaMath}, which use the training responses from GSM8K \citep{GSM8K} or MATH \citep{MATH}, our training dataset contains only their input prompts, making the results more generalizable. Furthermore, we do not use any modifications of the original question as an intermediate step, such as backward questions in \citet{MetaMath} or Evol-Instructions in \citet{WizardMath}, which might yield additional gains.

 We perform our distillation on two pairs of teacher and student: (1) Phi3-4B to TinyLlama, and (2) Phi3-14B to Llama2-7B \citep{Llama2}. We do not fine-tune the teachers on the dataset and assume them to be sufficiently capable in mathematical reasoning to produce supervision signals. For every pair of teacher and student, our distillation is performed in two stages, 
\begin{enumerate}[nosep]
    \item Pretraining distillation on 2.5B tokens from the OWM corpus ($\beta=2.0$)
    \item Three epochs of distillation on the ORCAMEL dataset for the same teacher--student pair.
\end{enumerate}

We also add a baseline by fine-tuning TinyLlama on the ORCAMEL dataset, after pretraining it on the same 2.5B OWM tokens without any distillation. The performance of the distilled models is comparable to that of Rho-1 \citep{Rho1}. Rho-1 is created by continuing TinyLlama's pretraining on 30B tokens from the OWM corpus, using reducible holdout (Rho) loss selection \citep{rholoss} to eliminate noisy tokens, achieving SOTA results on mathematical tasks with models of around 1B parameters.  The distilled Llama2-7B outperforms SOTA models for Maths inference built using Llama-2 as the base model, such as Llemma-7B \citep{Llemma}, Orca-2 \citep{ORCA-math}, or Wizard-Math \citep{WizardMath}, and we generated their results using the same Mathematical evaluation harness. Further, our method has a much lower compute budget than the next-best model, Llemma-7B, as explained in \Cref{app:llema}. Although unsupervised corpora for pretraining are unlimited, supervised datasets are always limited. It is better to use them with a teacher's supervision for optimal performance, rather than merely fine-tuning the student on them.

\section{Related Work}

Most of the work in KD for LLMs focuses on task-specific knowledge transfer via instruction prompts, following Sequence-KD \citep{SeqKD}, in which the teacher generates a sequence-specific prompt and the student is fine-tuned on that sequence. Recently, there has been a surge in reinforcement learning-based policy optimization for distillation, like MiniLLM and \citet{OnPolicyKD}. However, these methods involve generating sequences from the student during training, which can be expensive for large datasets. Recently, DistilLM \citep{DistilLM} addressed this issue by implementing an efficient generation scheduler. Overall, these on-policy methods are limited to small datasets; for example, both DistilLM and MiniLLM use the DollyEval dataset, which contains 15,000 data points. They cannot be applied to large-scale datasets exceeding 200K, which is standard for distillation in summarization or translation (\citet{Summary}, \citet{OnPolicyKD}).


When it comes to large-scale pretraining distillation to prepare the student from scratch, there is work on encoder-only models, such as DistilBERT \citep{distillbert} or MiniLM \citep{minilm}.  Work like \citet{Summary} extends it to encoder--decoder models for generative tasks such as summarization or machine translation. However, most pretraining distillation in causal models, such as distilling Gemma2 models from Gemini \citep{Gemma2} or work like \citet{KD_nVidia}, still follows logit matching with minimal modification. MiniPLM is the only work we found that attempts distillation without logit matching.

Works like MiniPLM, MiniLLM, or On-policy KD of \citet{OnPolicyKD} uses the reverse KL divergence instead of the forward one. However, the mode-seeking behavior of reverse KLD will suppress the contribution of words other than the one with the maximum probability. Furthermore, as shown in \Cref{tb:Larger}, reverse KL yields student models with the worst calibration, implying that reverse KL-based methodology is not suitable for unsupervised distillation in our settings.


\section{Conclusion}

Here, we present a novel distillation algorithm for language models that extends the commonly used KL divergence, and we demonstrate its competitiveness through extensive experiments. Works such as Sequence-KD and MiniLLM are not well-suited to pretraining on large-scale datasets. MiniPLM performs poorly for domain-specific distillation and cannot be directly applied to supervised tasks. In contrast, our method applies to both pretraining and supervised distillation, and is substantially cheaper for the latter because it requires neither teacher decoding (as in Seq-KD) nor student generation (as in MiniLLM or DistilLM \citep{DistilLM}). Consequently, TAD has a computational burden comparable to Vanilla KD, enabling large-scale pretraining distillation within a limited GPU budget. Finally, we show that it can be used to train competitive models for mathematical reasoning on publicly available datasets. Taken together with its modest computational requirements, TAD provides a compelling and versatile distillation method for causal LMs. 

\section{Acknowledgement}
The authors thank Dr. Lester Mackey for his valuable discussions on the methodology. This research was supported by The University of Melbourne’s Research Computing Services (Spartan) and the Petascale Campus Initiative.

\nocite{ACL2023}
\bibliography{example_paper}

@inproceedings{MiniPLM,
  title={{MiniPLM}: Knowledge Distillation for Pre-Training Language Models},
  author={Gu, Yuxian and Zhou, Hao and Meng, Fandong and Zhou, Jie and Huang, Minlie},
  booktitle={The Thirteenth International Conference on Learning Representations},
  year={2025}
}

@article{taylor2022galactica,
  title={Galactica: A large language model for science},
  author={Taylor, Ross and Kardas, Marcin and Cucurull, Guillem and Scialom, Thomas and Hartshorn, Anthony and Saravia, Elvis and Poulton, Andrew and Kerkez, Viktor and Stojnic, Robert},
  journal={arXiv preprint arXiv:2211.09085},
  year={2022}
}

@article{zhang2022opt,
  title={Opt: Open pre-trained transformer language models},
  author={Zhang, Susan and Roller, Stephen and Goyal, Naman and Artetxe, Mikel and Chen, Moya and Chen, Shuohui and Dewan, Christopher and Diab, Mona and Li, Xian and Lin, Xi Victoria and others},
  journal={arXiv preprint arXiv:2205.01068},
  year={2022}
}

@inproceedings{sparselogitsampling,
  title={Sparse Logit Sampling: Accelerating Knowledge Distillation in LLMs},
  author={Anshumann, Anshumann and Zaidi, Mohd Abbas and Kedia, Akhil and Ahn, Jinwoo and Kwon, Taehwak and Lee, Kangwook and Lee, Haejun and Lee, Joohyung},
  booktitle={Proceedings of the 63rd Annual Meeting of the Association for Computational Linguistics (Volume 1: Long Papers)},
  pages={18085--18108},
  year={2025}
}

@inproceedings{iwana2019explaining,
  title={Explaining convolutional neural networks using softmax gradient layer-wise relevance propagation},
  author={Iwana, Brian Kenji and Kuroki, Ryohei and Uchida, Seiichi},
  booktitle={2019 IEEE/CVF International Conference on Computer Vision Workshop (ICCVW)},
  pages={4176--4185},
  year={2019},
  organization={IEEE}
}

@inproceedings{stochasticbeam,
  title={Stochastic beams and where to find them: The gumbel-top-k trick for sampling sequences without replacement},
  author={Kool, Wouter and Van Hoof, Herke and Welling, Max},
  booktitle={International conference on machine learning},
  pages={3499--3508},
  year={2019},
  organization={PMLR}
}

@inproceedings{TopK_Bernt_Schiele,
  title={Loss functions for top-k error: Analysis and insights},
  author={Lapin, Maksim and Hein, Matthias and Schiele, Bernt},
  booktitle={Proceedings of the IEEE conference on computer vision and pattern recognition},
  pages={1468--1477},
  year={2016}
}

@inproceedings{CKA_ICLR,
  title={Improving language model distillation through hidden state matching},
  author={Dasgupta, Sayantan and Cohn, Trevor},
  booktitle={The Thirteenth International Conference on Learning Representations},
  year={2025}
}

@article{RelativeChange,
  title={How should relative changes be measured?},
  author={T{\"o}rnqvist, Leo and Vartia, Pentti and Vartia, Yrj{\"o} O},
  journal={The American Statistician},
  volume={39},
  number={1},
  pages={43--46},
  year={1985},
  publisher={Taylor \& Francis}
}

@article{ORCA-math,
  title={Orca-math: Unlocking the potential of slms in grade school math},
  author={Mitra, Arindam and Khanpour, Hamed and Rosset, Corby and Awadallah, Ahmed},
  journal={arXiv preprint arXiv:2402.14830},
  year={2024}
}

@inproceedings{rholoss,
  title={Prioritized training on points that are learnable, worth learning, and not yet learnt},
  author={Mindermann, S{\"o}ren and Brauner, Jan M and Razzak, Muhammed T and Sharma, Mrinank and Kirsch, Andreas and Xu, Winnie and H{\"o}ltgen, Benedikt and Gomez, Aidan N and Morisot, Adrien and Farquhar, Sebastian and others},
  booktitle={International Conference on Machine Learning},
  pages={15630--15649},
  year={2022},
  organization={PMLR}
}

@article{WizardMath,
  title={Wizardmath: Empowering mathematical reasoning for large language models via reinforced evol-instruct},
  author={Luo, Haipeng and Sun, Qingfeng and Xu, Can and Zhao, Pu and Lou, Jianguang and Tao, Chongyang and Geng, Xiubo and Lin, Qingwei and Chen, Shifeng and Zhang, Dongmei},
  journal={arXiv preprint arXiv:2308.09583},
  year={2023}
}

@article{MetaMath,
  title={Metamath: Bootstrap your own mathematical questions for large language models},
  author={Yu, Longhui and Jiang, Weisen and Shi, Han and Yu, Jincheng and Liu, Zhengying and Zhang, Yu and Kwok, James T and Li, Zhenguo and Weller, Adrian and Liu, Weiyang},
  journal={arXiv preprint arXiv:2309.12284},
  year={2023}
}

@article{MATH,
  title={Solving quantitative reasoning problems with language models},
  author={Lewkowycz, Aitor and Andreassen, Anders and Dohan, David and Dyer, Ethan and Michalewski, Henryk and Ramasesh, Vinay and Slone, Ambrose and Anil, Cem and Schlag, Imanol and Gutman-Solo, Theo and others},
  journal={Advances in Neural Information Processing Systems},
  volume={35},
  pages={3843--3857},
  year={2022}
}

@article{GSM8K,
  title={Training Verifiers to Solve Math Word Problems},
  author={Cobbe, Karl and Kosaraju, Vineet and Bavarian, Mohammad and Chen, Mark and Jun, Heewoo and Kaiser, Lukasz and Plappert, Matthias and Tworek, Jerry and Hilton, Jacob and Nakano, Reiichiro and Hesse, Christopher and Schulman, John},
  journal={arXiv preprint arXiv:2110.14168},
  year={2021}
}

@misc{camel-math,
      title={CAMEL: Communicative Agents for "Mind" Exploration of Large Scale Language Model Society}, 
      author={Guohao Li and Hasan Abed Al Kader Hammoud and Hani Itani and Dmitrii Khizbullin and Bernard Ghanem},
      year={2023},
      eprint={2303.17760},
      archivePrefix={arXiv},
      primaryClass={cs.AI}
}

@article{OpenWebMath,
  title={Openwebmath: An open dataset of high-quality mathematical web text},
  author={Paster, Keiran and Santos, Marco Dos and Azerbayev, Zhangir and Ba, Jimmy},
  journal={arXiv preprint arXiv:2310.06786},
  year={2023}
}

@article{Adam,
  title={Adam: A method for stochastic optimization},
  author={Kingma, Diederik P and Ba, Jimmy},
  journal={arXiv preprint arXiv:1412.6980},
  year={2014}
}

@inproceedings{DistilLM,
  title={DistiLLM: Towards Streamlined Distillation for Large Language Models},
  author={Ko, Jongwoo and Kim, Sungnyun and Chen, Tianyi and Yun, Se-Young},
  booktitle={International Conference on Machine Learning},
  pages={24872--24895},
  year={2024},
  organization={PMLR}
}

@article{PPO,
  title={Proximal policy optimization algorithms},
  author={Schulman, John and Wolski, Filip and Dhariwal, Prafulla and Radford, Alec and Klimov, Oleg},
  journal={arXiv preprint arXiv:1707.06347},
  year={2017}
}

@article{Llemma,
  title={Llemma: An open language model for mathematics},
  author={Azerbayev, Zhangir and Schoelkopf, Hailey and Paster, Keiran and Santos, Marco Dos and McAleer, Stephen and Jiang, Albert Q and Deng, Jia and Biderman, Stella and Welleck, Sean},
  journal={arXiv preprint arXiv:2310.10631},
  year={2023}
}

@article{Rho1,
  title={Rho-1: Not all tokens are what you need},
  author={Lin, Zhenghao and Gou, Zhibin and Gong, Yeyun and Liu, Xiao and Shen, Yelong and Xu, Ruochen and Lin, Chen and Yang, Yujiu and Jiao, Jian and Duan, Nan and others},
  journal={arXiv preprint arXiv:2404.07965},
  year={2024}
}

@article{CoT,
  title={Chain-of-thought prompting elicits reasoning in large language models},
  author={Wei, Jason and Wang, Xuezhi and Schuurmans, Dale and Bosma, Maarten and Xia, Fei and Chi, Ed and Le, Quoc V and Zhou, Denny and others},
  journal={Advances in neural information processing systems},
  volume={35},
  pages={24824--24837},
  year={2022}
}

@article{Llama2,
  title={Llama 2: Open foundation and fine-tuned chat models},
  author={Touvron, Hugo and Martin, Louis and Stone, Kevin and Albert, Peter and Almahairi, Amjad and Babaei, Yasmine and Bashlykov, Nikolay and Batra, Soumya and Bhargava, Prajjwal and Bhosale, Shruti and others},
  journal={arXiv preprint arXiv:2307.09288},
  year={2023}
}

@inproceedings{ECE,
  title={On calibration of modern neural networks},
  author={Guo, Chuan and Pleiss, Geoff and Sun, Yu and Weinberger, Kilian Q},
  booktitle={International conference on machine learning},
  pages={1321--1330},
  year={2017},
  organization={PMLR}
}

@article{fullECE,
  title={Full-ECE: A Metric For Token-level Calibration on Large Language Models},
  author={Liu, Han and Zhang, Yupeng and Wang, Bingning and Chen, Weipeng and Hu, Xiaolin},
  journal={arXiv preprint arXiv:2406.11345},
  year={2024}
}

@inproceedings{ACL2023,
    title = "Cost-effective Distillation of Large Language Models",
    author = "Dasgupta, Sayantan  and
      Cohn, Trevor  and
      Baldwin, Timothy",
    editor = "Rogers, Anna  and
      Boyd-Graber, Jordan  and
      Okazaki, Naoaki",
    booktitle = "Findings of the Association for Computational Linguistics: ACL 2023",
    month = jul,
    year = "2023",
    address = "Toronto, Canada",
    publisher = "Association for Computational Linguistics",
    url = "https://aclanthology.org/2023.findings-acl.463/",
    doi = "10.18653/v1/2023.findings-acl.463",
    pages = "7346--7354"
}

@misc{LMEH,
  author       = {Gao, Leo and Tow, Jonathan and Abbasi, Baber and Biderman, Stella and Black, Sid and DiPofi, Anthony and Foster, Charles and Golding, Laurence and Hsu, Jeffrey and Le Noac'h, Alain and Li, Haonan and McDonell, Kyle and Muennighoff, Niklas and Ociepa, Chris and Phang, Jason and Reynolds, Laria and Schoelkopf, Hailey and Skowron, Aviya and Sutawika, Lintang and Tang, Eric and Thite, Anish and Wang, Ben and Wang, Kevin and Zou, Andy},
  title        = {A framework for few-shot language model evaluation},
  month        = 07,
  year         = 2024,
  publisher    = {Zenodo},
  version      = {v0.4.3},
  doi          = {10.5281/zenodo.12608602},
  url          = {https://zenodo.org/records/12608602}
}

@article{KD_nVidia,
  title={Compact Language Models via Pruning and Knowledge Distillation},
  author={Muralidharan, Saurav and Sreenivas, Sharath Turuvekere and Joshi, Raviraj and Chochowski, Marcin and Patwary, Mostofa and Shoeybi, Mohammad and Catanzaro, Bryan and Kautz, Jan and Molchanov, Pavlo},
  journal={arXiv preprint arXiv:2407.14679},
  year={2024}
}

@article{flashattention,
  title={Flashattention: Fast and memory-efficient exact attention with io-awareness},
  author={Dao, Tri and Fu, Dan and Ermon, Stefano and Rudra, Atri and R{\'e}, Christopher},
  journal={Advances in Neural Information Processing Systems},
  volume={35},
  pages={16344--16359},
  year={2022}
}

@article{minilm,
  title={Minilm: Deep self-attention distillation for task-agnostic compression of pre-trained transformers},
  author={Wang, Wenhui and Wei, Furu and Dong, Li and Bao, Hangbo and Yang, Nan and Zhou, Ming},
  journal={Advances in Neural Information Processing Systems},
  volume={33},
  pages={5776--5788},
  year={2020}
}

@article{SeqKD,
  title={Sequence-level knowledge distillation},
  author={Kim, Yoon and Rush, Alexander M},
  journal={arXiv preprint arXiv:1606.07947},
  year={2016}
}

@article{Huggingface,
  author    = {Thomas Wolf and
               Lysandre Debut and
               Victor Sanh and
               Julien Chaumond and
               Clement Delangue and
               Anthony Moi and
               Pierric Cistac and
               Tim Rault and
               R{\'{e}}mi Louf and
               Morgan Funtowicz and
               Jamie Brew},
  title     = {{HuggingFace's} Transformers: State-of-the-art Natural Language Processing},
  journal   = {CoRR},
  volume    = {abs/1910.03771},
  year      = {2019},
  url       = {http://arxiv.org/abs/1910.03771},
  bibsource = {dblp computer science bibliography, https://dblp.org}
}

@article{MBART,
  title={Multilingual translation with extensible multilingual pretraining and finetuning},
  author={Tang, Yuqing and Tran, Chau and Li, Xian and Chen, Peng-Jen and Goyal, Naman and Chaudhary, Vishrav and Gu, Jiatao and Fan, Angela},
  journal={arXiv preprint arXiv:2008.00401},
  year={2020}
}

@article{Summary,
  title={Pre-trained summarization distillation},
  author={Shleifer, Sam and Rush, Alexander M},
  journal={arXiv preprint arXiv:2010.13002},
  year={2020}
}

@inproceedings{MiniLLM,
  title={MiniLLM: Knowledge distillation of large language models},
  author={Gu, Yuxian and Dong, Li and Wei, Furu and Huang, Minlie},
  booktitle={The Twelfth International Conference on Learning Representations},
  year={2024}
}

@inproceedings{Multi-Teacher,
  title={One Teacher is Enough? Pre-trained Language Model Distillation from Multiple Teachers},
  author={Wu, Chuhan and Wu, Fangzhao and Huang, Yongfeng},
  booktitle={Findings of the Association for Computational Linguistics: ACL-IJCNLP 2021},
  pages={4408--4413},
  year={2021}
}

@article{Pile,
  title={The pile: An 800gb dataset of diverse text for language modeling},
  author={Gao, Leo and Biderman, Stella and Black, Sid and Golding, Laurence and Hoppe, Travis and Foster, Charles and Phang, Jason and He, Horace and Thite, Anish and Nabeshima, Noa and others},
  journal={arXiv preprint arXiv:2101.00027},
  year={2020}
}

@article{distillbert,
  author    = {Victor Sanh and
               Lysandre Debut and
               Julien Chaumond and
               Thomas Wolf},
  title     = {{DistilBERT}, a distilled version of {BERT:} smaller, faster, cheaper
               and lighter},
  journal   = {CoRR},
  volume    = {abs/1910.01108},
  year      = {2019},
  url       = {http://arxiv.org/abs/1910.01108},
  eprinttype = {arXiv},
  eprint    = {1910.01108},
  bibsource = {dblp computer science bibliography, https://dblp.org}
}

@inproceedings{KD_Hinton,
  doi = {10.48550/ARXIV.1503.02531},
  
  url = {https://arxiv.org/abs/1503.02531},
  
  author = {Hinton, Geoffrey and Vinyals, Oriol and Dean, Jeff},
  
  title = {Distilling the Knowledge in a Neural Network},
  
  booktitle = {NIPS 2014 Deep Learning Workshop},
  
  year = {2014},
  
  copyright = {arXiv.org perpetual, non-exclusive license}
}

@inproceedings{OnPolicyKD,
  title={On-policy distillation of language models: Learning from self-generated mistakes},
  author={Agarwal, Rishabh and Vieillard, Nino and Zhou, Yongchao and Stanczyk, Piotr and Garea, Sabela Ramos and Geist, Matthieu and Bachem, Olivier},
  booktitle={The Twelfth International Conference on Learning Representations},
  year={2024}
}

@article{Phi2,
  title={Phi-2: The surprising power of small language models},
  author={Javaheripi, Mojan and Bubeck, S{\'e}bastien and Abdin, Marah and Aneja, Jyoti and Bubeck, Sebastien and Mendes, Caio C{\'e}sar Teodoro and Chen, Weizhu and Del Giorno, Allie and Eldan, Ronen and Gopi, Sivakanth and others},
  journal={Microsoft Research Blog},
  volume={1},
  number={3},
  pages={3},
  year={2023}
}

@inproceedings{DKD,
  title={Decoupled knowledge distillation},
  author={Zhao, Borui and Cui, Quan and Song, Renjie and Qiu, Yiyu and Liang, Jiajun},
  booktitle={Proceedings of the IEEE/CVF Conference on computer vision and pattern recognition},
  pages={11953--11962},
  year={2022}
}

@article{Regmix,
  title={Regmix: Data mixture as regression for language model pre-training},
  year = {2024},
  author={Liu, Qian and Zheng, Xiaosen and Muennighoff, Niklas and Zeng, Guangtao and Dou, Longxu and Pang, Tianyu and Jiang, Jing and Lin, Min},
  journal={arXiv preprint arXiv:2407.01492},
}

@article{Qwen2.5,
  title={Qwen2. 5 technical report},
  author={Yang, An and Yang, Baosong and Zhang, Beichen and Hui, Binyuan and Zheng, Bo and Yu, Bowen and Li, Chengyuan and Liu, Dayiheng and Huang, Fei and Wei, Haoran and others},
  journal={arXiv preprint arXiv:2412.15115},
  year={2024}
}

@article{Gemma2,
  title={Gemma 2: Improving open language models at a practical size},
  author={Team, Gemma and Riviere, Morgane and Pathak, Shreya and Sessa, Pier Giuseppe and Hardin, Cassidy and Bhupatiraju, Surya and Hussenot, L{\'e}onard and Mesnard, Thomas and Shahriari, Bobak and Ram{\'e}, Alexandre and others},
  journal={arXiv preprint arXiv:2408.00118},
  year={2024}
}
\bibliographystyle{icml2026}

\appendix

\section{Derivation of the Gradient}
\label{app:derivation}

Here we present an elaborated derivation of the gradients. The derivations follow the material in the appendix of \citet{sparselogitsampling}. If $p_i = \exp(z_i)/\sum_{i=1}^{|\mathcal{V}|}\exp(z_i)$ is the softmax probability for a logit $z_i$ for a vocabulary $\mathcal{V}$, then the gradient of $p_k$ is (from \citep{iwana2019explaining}):
\begin{equation}
\label{eqn:derv-softmax}
    \frac{\partial p_j}{\partial z_i} = p_j\left(\mathds{1}_{[i=j]}-p_i\right)
\end{equation}

Now, given a vocabulary $\mathcal{V}$, the KL Divergence loss between the teacher probabilities of the teacher ($p^T_i $) and the student ($p^S_i $) is:
\begin{equation}
    \mathcal{L}_{KLD} = \sum_{i=1}^{|\mathcal{V}|} p^T_i \log (p^T_i/p^S_i)
\end{equation}

It can be derived that, 
\begin{align*}
       \frac{\partial \mathcal{L}_{KLD}}{\partial z_i} &=  -\sum_{j=1}^{|\mathcal{V}|} \frac{p^T_j}{p^S_j}\frac{\partial p^S_j}{\partial z_i} \\
       &=  -\sum_{j=1}^{|\mathcal{V}|} p^T_j \left(\mathds{1}_{[i=j]}-p^S_i\right) \\
       & =p^S_i \cdot (\sum_{j=1}^{|\mathcal{V}|}p^T_j)  -\sum_{j=1}^{|\mathcal{V}|} p^T_j\mathds{1}_{[i=j]}  \\
       &= p^S_i - p^T_i
\label{eqn:kld-grad} 
\numberthis
\end{align*}

Now, we can show that $\mathcal{\mathcal{D}}_{KL_1}$ has $K+1$ terms when we consider top-$K$ probabilities, with the first $K$ being ($i \in [K]$) 
\begin{align*}
    L_{1:K} = \sum_{k=1}^K \accentset{\ast}{p}^T_k \log \frac{\accentset{\ast}{p}^T_k}{\accentset{\ast}{p}^S_k}
\end{align*}

\noindent where $\accentset{\ast}{p}^S_k$ are the student probabilities corresponding to the top-$K$ tokens, i.e. tokens for which the teacher probabilities are maximum. The derivative of $L_{1:K}$ w.r.t.\ a logit $z_i$ is 
\begin{equation}
    \frac{\partial L_{1:K}}{\partial z_i} = p^S_i \cdot (\sum_{k=1}^{K} \accentset{\ast}{p}^T_k)  -\sum_{k=1}^{K} \accentset{\ast}{p}^T_k\mathds{1}_{[i=k]} 
\end{equation}

Now for $i \in [\mathcal{V} \setminus K]$, the indicator function $\mathds{1}_{[i=k]}$ is never one. Therefore, the gradient of $L_{1:K}$ has the following forms for two different cases, as:
\begin{align*}
    \frac{\partial L_{1:K}}{\partial z_i} =
    \begin{cases} p^S_i \cdot (\sum_{k=1}^{K}\accentset{\ast}{p}^T_i) - p^T_i \qquad &  i \in [K] \\
    p^S_i \cdot (\sum_{k=1}^{K}\accentset{\ast}{p}^T_i) &i \in [\mathcal{V} \setminus K]
    \end{cases}
\end{align*}

\noindent  Please note that the top $K$ probabilities do not sum to one. The last term $L_{K+1}$ can be expressed as:
\begin{align*}
    L_{K+1} &= \left(1-\sum_{i=1}^K \accentset{\ast}{p}^T_k\right) \log \frac{1-\sum_{i=1}^K \accentset{\ast}{p}^T_k}{1-\sum_{i=1}^K \accentset{\ast}{p}^S_i} \\
    &= - \left(1-\sum_{k=1}^K \accentset{\ast}{p}^T_k\right)\cdot \log {\left(1-\sum_{k=1}^K \accentset{\ast}{p}^S_k\right)} + \text{C} 
\end{align*}

\noindent where C is a constant. The derivative of the last term, using the derivative of $p^S_k$ from \Cref{eqn:derv-softmax} is:
\begin{align*}
    \frac{\partial L_{K+1}}{\partial z_i} & 
    =\frac{1-\sum_{k=1}^K \accentset{\ast}{p}^T_k}{1-\sum_{k=1}^K \accentset{\ast}{p}^S_k} \cdot \sum_{k=1}^K \frac{\partial \accentset{\ast}{p}^S_k}{\partial z_i} \\
    &= \frac{1-\sum_{k=1}^K \accentset{\ast}{p}^T_k}{1-\sum_{k=1}^K \accentset{\ast}{p}^S_k} \cdot \sum_{k=1}^K \accentset{\ast}{p}^S_k\left(\mathds{1}_{[i=k]}-p^S_i\right) \\
\end{align*}

\begin{table*}[t]
\begin{center}
\begin{tabular}{crcccccc}  
 Teacher & \#P(M) & $|\mathcal{V}|$ & $d_S$ & $L_S$ & $n_H$ & $d_H$& $d_{FFN}$  \\ \toprule
 Qwen1.5-1.8B & $1.2$B & 151,936& 1,536& 24 & 16 & 96 & 4,224\\ 
 Qwen1.5-1.8B & $0.5$B & 151,936& 1,024& 24 & 16 & 64 & 2,816\\ 
 Phi2-2.8B & 1.1B & 52,000 & 2,560 & 16 & 32 & 80 & 5,120 \\
 Qwen2.5-3B & 1.2B & 151,936 & 2,048 & 18 & 16 & 128 & 7,680 \\
 Gemma2-9B & 2B & 256,000 & 3,584 & 14 & 16 & 224 & 4,096\\
 \end{tabular}
\caption{The architectures of different students used in distillation for pretraining from scratch}
\label{tb:Architecture}
\end{center}
\end{table*}

Again, for $i \in [\mathcal{V} \setminus K]$, the indicator function $\mathds{1}_{[i=k]}$ is never one. Therefore,
\begin{equation}
  \frac{\partial L_{K+1}}{\partial z_i}=
  \begin{cases}
p^S_i \cdot \left(1-\sum_{k=1}^K \accentset{\ast}{p}^T_k \right) &  i \in [K]\\
-p^S_i \cdot \left(\frac{1-\sum_{k=1}^K \accentset{\ast}{p}^T_k}{1-\sum_{k=1}^K \accentset{\ast}{p}^S_k} \right)   \sum_{k=1}^K \accentset{\ast}{p}^T_k & i \in [\mathcal{V\setminus K}]
\end{cases}  
\end{equation}

Combining the gradients of $L_{1:K}$ and $L_{K+1}$, since $\mathcal{\mathcal{D}}_{KL_1} = L_{1:K} + L_{K+1}$

\begin{equation}
  \frac{\partial \mathcal{\mathcal{D}}_{KL_1}}{\partial z_i}=
  \begin{cases}
p^S_i - p^T_i  &  i \in [K]\\
p^S_i \cdot \left(\frac{\sum_{k=1}^K \accentset{\ast}{p}^T_k-\sum_{k=1}^K \accentset{\ast}{p}^S_k}{1-\sum_{k=1}^K \accentset{\ast}{p}^S_k} \right)  & i \in [\mathcal{V\setminus K}]
\end{cases}  
\label{eqn:grad-dkl1}
\end{equation}

Therefore, the gradients of the logits corresponding to the tokens of top-$K$ teacher probabilities remain the same, while the gradients of the logits corresponding to the rest of the tokens change. The second term  $\mathcal{\mathcal{D}}_{KL_2}$ solely depends on the logits of the rest of the tokens.

\begin{equation}
    \mathcal{\mathcal{D}}_{KL_2} = \sum_{i \in \mathcal{V} \setminus K} \tilde{p}^T_i \log \frac{\tilde{p}^T_i}{\tilde{p}^S_i}
\end{equation}

\noindent where we can generate $\tilde{p}^S_i$ directly from $z_i$ as $\tilde{p}^S_i = \frac{\exp{z_i}}{\sum_{k \in \mathcal{V} \setminus K} \exp{z_k}}$. Also, $\tilde{p}^T_i$ comes from a similar softmax, but is constant. Therefore,

\begin{align*}
    \frac{\partial {\mathcal{D}}_{KL_2}}{\partial z_i} =
    \begin{cases} 
    0 &i \in [K] \\
     \tilde{p}^S_i - \tilde{p}^T_i & i \in [\mathcal{V\setminus K}]
    \end{cases}
\end{align*}

The gradients of the logits of the top-$K$ tokens are zero for ${\mathcal{D}}_{KL_2}$; only their gradient for ${\mathcal{D}}_{KL_1}$ is non-zero (\Cref{eqn:grad-dkl1}). And as a result, their gradient is the same as that for ordinary KL Divergence (\Cref{eqn:kld-grad}). Therefore, Decoupled KD does \textbf{not} change the gradient of the logits of the top-$K$ tokens.

As for the logits of the non-top-$K$ tokens, their gradient for ${\mathcal{D}}_{KL_2}$ can be written as,
\begin{equation}
    \frac{\partial {\mathcal{D}}_{KL_2}}{\partial z_i} = \frac{p^S_i}{1-\sum_{k=1}^K \accentset{\ast}{p}^S_k} - \frac{p^T_i}{1-\sum_{k=1}^K \accentset{\ast}{p}^T_k}
\end{equation}
\noindent since $\tilde{p}^T_i =  \frac{p^T_i}{1-\sum_{k=1}^K \accentset{\ast}{p}^T_k}$ and $\tilde{p}^S_i =  \frac{p^S_i}{1-\sum_{k=1}^K \accentset{\ast}{p}^S_k}$

Therefore,
\begin{equation}
\label{eqn:grad-dkl2}
    \left(1-\sum_{k=1}^K \accentset{\ast}{p}^T_k \right)\frac{\partial {\mathcal{D}}_{KL_2}} {\partial z_i}
    = p^S_i\cdot \frac{1-\sum_{k=1}^K \accentset{\ast}{p}^T_k}{1-\sum_{k=1}^K \accentset{\ast}{p}^S_k} - p^T_i
\end{equation}

Combining the derivative of ${\mathcal{D}}_{KL_2}$ from (\Cref{eqn:grad-dkl1} for the tail logits, i.e., for $i \in [\mathcal{V\setminus K}]$, it can easily be checked that
\begin{align*}
    \small
    &\frac{\partial {\mathcal{D}}_{KL_1}} {\partial z_i} + \left(1-\sum_{k=1}^K \accentset{\ast}{p}^T_k \right)\frac{\partial {\mathcal{D}}_{KL_2}} {\partial z_i} \\
   & = \left(\frac{p^S_i \cdot \sum_{k=1}^K \accentset{\ast}{p}^T_k-p^S_i \cdot \sum_{k=1}^K \accentset{\ast}{p}^S_k  }{1-\sum_{k=1}^K \accentset{\ast}{p}^S_k} \right) \\
   & \phantom{AAAAAAAAAA} + \left(\frac{ p^S_i -p^S_i \cdot \sum_{k=1}^K \accentset{\ast}{p}^T_k}{1-\sum_{k=1}^K \accentset{\ast}{p}^S_k} \right) - p^T_i \\
   & = p^S_i - p^T_i
\end{align*}
Since $\mathcal{L}_{KLD} = {\mathcal{D}}_{KL_1} + \left(1-\sum_{k=1}^K \accentset{\ast}{p}^T_k \right){\mathcal{D}}_{KL_2}$, their gradients are the same. Now, for Decoupled KD, the divergence is:  $\mathcal{L}_{DIV} = {\mathcal{D}}_{KL_1} + \beta(X)\left(1-\sum_{k=1}^K \accentset{\ast}{p}^T_k \right){\mathcal{D}}_{KL_2}$, where $\beta(X) =  \beta/ (\frac{1}{N}\sum_{t=1}^{N} (1-\sum_{k=1}^K \accentset{\ast}{p}^T_{k}(t)))$, where $t$ is the index of a token in a sequence $X$ containing a total of $N$ tokens. This also means,
\begin{align*}
    \mathcal{L}_{DIV} &= {\mathcal{D}}_{KL_1} +  \left(1-\sum_{k=1}^K \accentset{\ast}{p}^T_k \right){\mathcal{D}}_{KL_2}\\
    & \phantom{AAAAAA} (\beta(X)-1)\left(1-\sum_{k=1}^K \accentset{\ast}{p}^T_k \right){\mathcal{D}}_{KL_2} \\
    &= \mathcal{L}_{KLD} + (\beta(X)-1)\left(1-\sum_{k=1}^K \accentset{\ast}{p}^T_k \right){\mathcal{D}}_{KL_2}
\end{align*}
    
Using \Cref{eqn:grad-dkl2}, the gradient of $\mathcal{L}_{DIV}$ has the following form for the logits $z_i$ for the tail tokens ($i \in [\mathcal{V}\setminus K]$)
\begin{align*}
    &\frac{\partial \mathcal{L}_{DIV}} {\partial z_i}\\
    &= \frac{\partial \mathcal{L}_{KLD}} {\partial z_i} 
 + (\beta(X)-1)\left(1-\sum_{k=1}^K \accentset{\ast}{p}^T_k \right)\frac{\partial {\mathcal{D}}_{KL_2}} {\partial z_i}\\
 & = p^S_i - p^T_i \\
 & \phantom{AAAA} + (\beta(X)-1) \left(p^S_i\cdot \frac{1-\sum_{k=1}^K \accentset{\ast}{p}^T_k}{1-\sum_{k=1}^K \accentset{\ast}{p}^S_k} - p^T_i\right)
\end{align*}

For the logits of the top-$K$ tokens, $\frac{\partial \mathcal{D}_{KL_2}} {\partial z_i}=0$, and therefore, their gradients are the same as those of Vanilla KD. This completes the derivation of the gradient of $\mathcal{L}_{DIV}$.

\bigskip

\section{Experimental Detail}
\label{app:experiments}

The architectures of different students for the pretraining from scratch are listed in \Cref{tb:Architecture}. All students have approximately $1$B active parameters, except for the $0.5$B student of Qwen, which has approximately $475$M active parameters. The architectures of the students of Qwen$1.5-1.8$B are kept the same as in the MiniPLM paper \citep{MiniPLM}.

The experiments are divided into two major parts: pretraining distillation from scratch, and continued pretraining. For pretraining distillation from scratch, we distilled the Qwen1.5, Phi2, and Qwen2.5 models on a single H100 GPU for a week, whereas we used 2 H100 GPUs for distilling the Gemma2-9B model. We used flash attention \citep{flashattention} whenever possible to speed up the computation, except for Gemma2. We used Adam optimizer \citep{Adam} with a learning rate of $\eta = 1e-4$ and a weight decay of $\lambda_d = 0.1$ for all the experiments. We used a batch size of $128$ for all the experiments.

For the continued pretraining distillation of Tiny-Llama, we used the Adam optimizer \citep{Adam} with a learning rate of $\eta = 3e-5$ and a weight decay of $\lambda_d = 0.1$ for all experiments. All experiments used a batch size of 128 and were conducted on a single NVIDIA H100 GPU. Supervised distillation is performed with a batch size of 32, $\eta = 1e-5$, $\lambda_d = 0.1$, and a context size of 2048.

\subsection{Cost of Supervised Distillation}
\label{app:llema}

 We conduct a comparative cost analysis of GPU hours required to produce state-of-the-art mathematical reasoning, starting with foundational models such as TinyLlama-1.1B and Llama2-7B. Models like Llemma or Rho-1 are trained using industrial resources. Rho-1 is trained for approximately 10 hours on a 32-GPU H100 stack, requiring a total of 320 GPU hours. The best model built on Llama-7B is Llemma, which was trained on A100 GPUs for 23,000 GPU hours. Even though it uses different hardware, we can establish an equivalence by noting that the 7B model in \citet{Rho1} takes the same number of GPU hours to train on an H100. It required 18 hours to train on 15 billion tokens using 32 H100 GPUs. Using their configuration setting, Llemma-7B will take 7,680 GPU hours to train on a single H100. This provides a reasonable estimate, since A100s are approximately a third slower than H100 GPUs for training ($ 23K \approx 3 \times 7,680$). Our two-stage method requires approximately 130 hours on a single H100 GPU for TinyLlama and 420 hours on two H100 GPUs (totaling 840 hours) for Llama-2, which is substantially cheaper than the existing methods.

\end{document}